\documentclass{article}

\usepackage[preprint]{neurips_2024}
\usepackage{authblk}
\usepackage{threeparttable}
\usepackage{graphicx}
\usepackage{amsmath} 
\usepackage{algorithm}
\usepackage{algorithmic}
\usepackage{adjustbox}
\usepackage{multirow}
\usepackage{float}
\usepackage[table]{xcolor}
\usepackage{caption}
\usepackage{subcaption}
\usepackage{tikz}
\definecolor{linegray}{RGB}{175, 171, 171}
\definecolor{lineblue}{RGB}{0, 176, 240}
\definecolor{linered}{RGB}{197, 90, 17}
\definecolor{T}{RGB}{146, 208, 80}
\definecolor{S}{RGB}{255, 217, 102}
\usepackage[utf8]{inputenc} 
\usepackage[T1]{fontenc}    
\usepackage{hyperref}       
\usepackage{url}            
\usepackage{booktabs}       
\usepackage{amsfonts}       
\usepackage{nicefrac}       
\usepackage{microtype}      
\usepackage{xcolor}         
\usepackage{pifont}

\usepackage[utf8]{inputenc} 
\usepackage[T1]{fontenc}    
\usepackage{hyperref}       
\usepackage{url}            
\usepackage{booktabs}       
\usepackage{amsfonts}       
\usepackage{nicefrac}       
\usepackage{microtype}      
\usepackage{xcolor}         
\usepackage{pifont}

\newcommand{\cmark}{\textcolor{green}{\ding{51}}}%
\newcommand{\xmark}{\textcolor{gray}{\ding{55}}}%

\title{SynRS3D\includegraphics[width=0.05\textwidth]{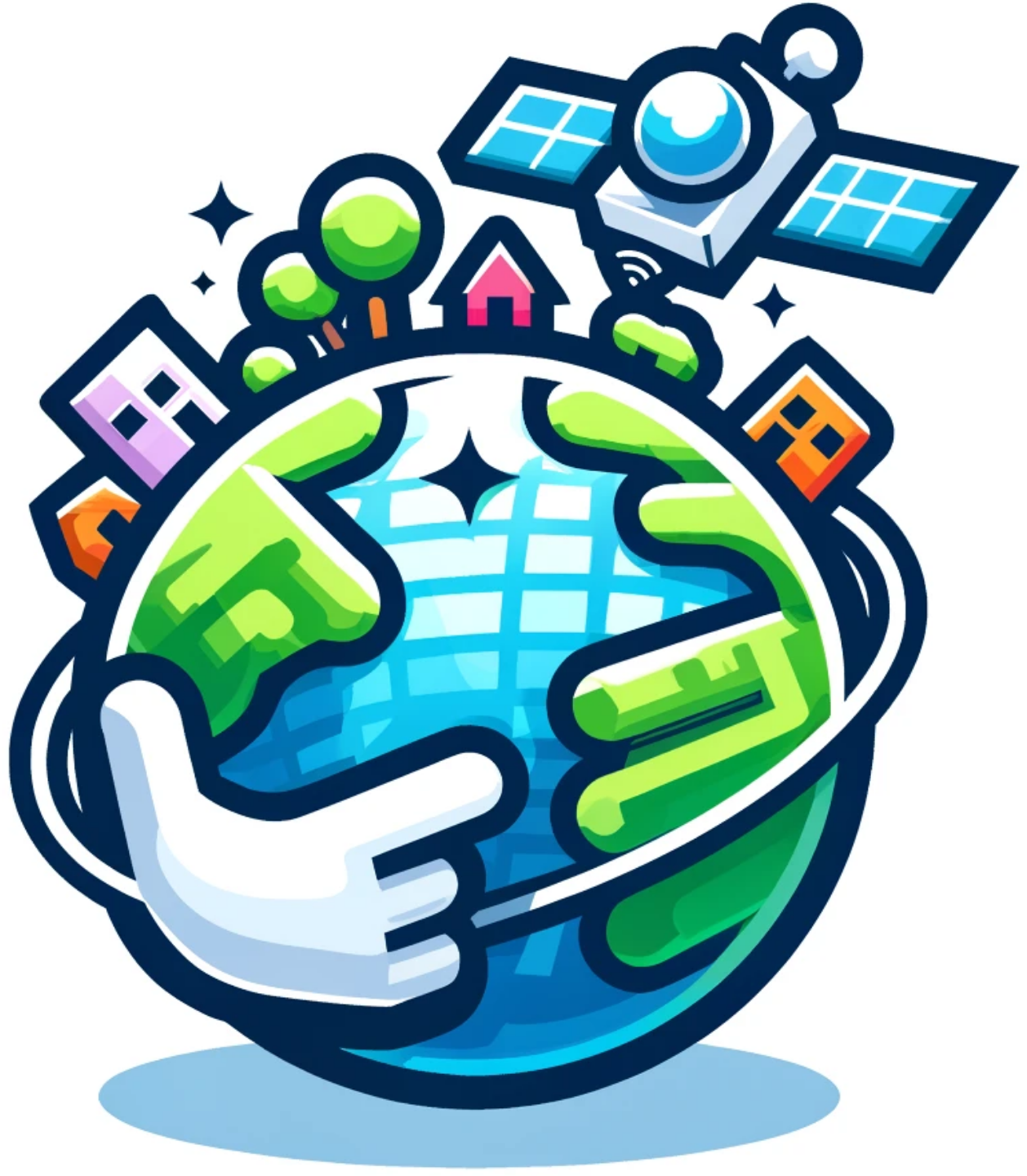}: A Synthetic Dataset for Global 3D Semantic Understanding from Monocular Remote Sensing Imagery}

\author[1,2]{\textbf{Jian Song}}
\author[1]{\textbf{Hongruixuan Chen}}
\author[1,2]{\textbf{Weihao Xuan}}
\author[2]{\textbf{Junshi Xia}}
\author[1,2]{\textbf{Naoto Yokoya}}

\affil[1]{The University of Tokyo, Tokyo, Japan}
\affil[2]{RIKEN AIP, Tokyo, Japan}

\begin{document}

\maketitle

\begin{abstract}
Global semantic 3D understanding from single-view high-resolution remote sensing (RS) imagery is crucial for Earth Observation (EO). However, this task faces significant challenges due to the high costs of annotations and data collection, as well as geographically restricted data availability. To address these challenges, synthetic data offer a promising solution by being easily accessible and thus enabling the provision of large and diverse datasets. We develop a specialized synthetic data generation pipeline for EO and introduce \textit{SynRS3D}, the largest synthetic RS 3D dataset. SynRS3D comprises 69,667 high-resolution optical images that cover six different city styles worldwide and feature eight land cover types, precise height information, and building change masks. To further enhance its utility, we develop a novel multi-task unsupervised domain adaptation (UDA) method, \textit{RS3DAda}, coupled with our synthetic dataset, which facilitates the RS-specific transition from synthetic to real scenarios for land cover mapping and height estimation tasks, ultimately enabling global monocular 3D semantic understanding based on synthetic data. Extensive experiments on various real-world datasets demonstrate the adaptability and effectiveness of our synthetic dataset and proposed RS3DAda method. SynRS3D and related codes will be available at \url{https://github.com/JTRNEO/SynRS3D}.
\end{abstract}

\begin{center}
\begin{figure*}[h]
  \centering
  \includegraphics[width=\textwidth]{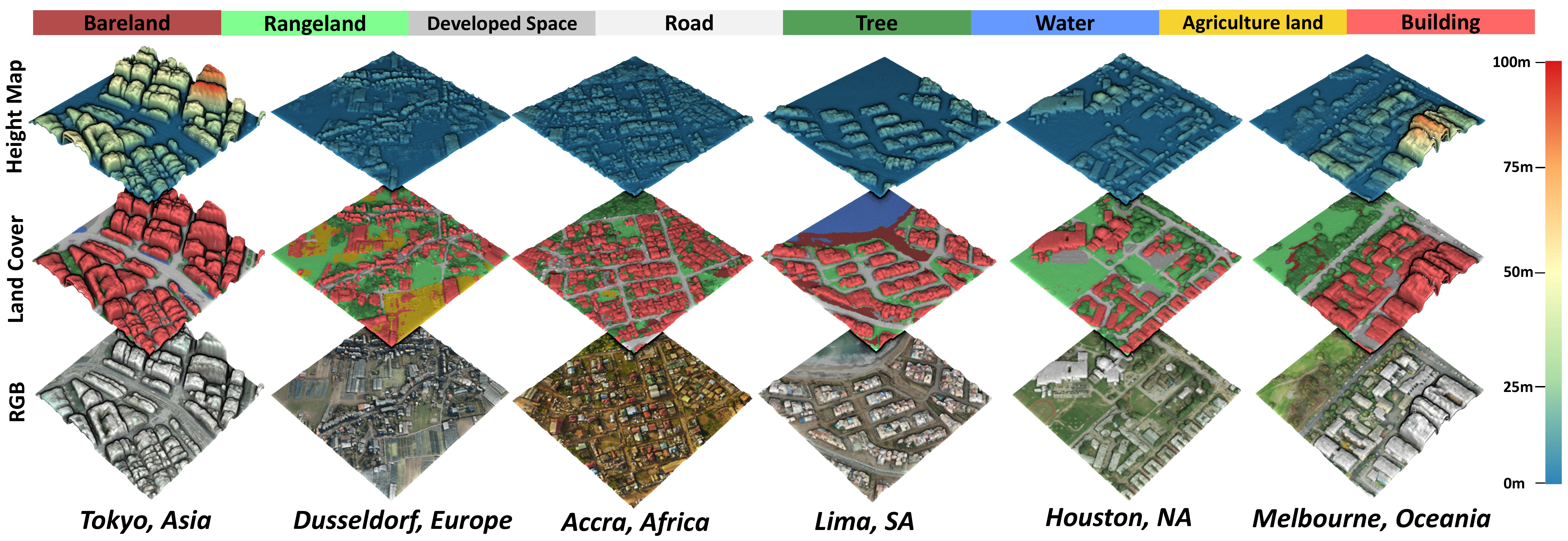}
  \caption{3D visualization outcomes from real-world monocular RS images, which uses the model trained on SynRS3D dataset with proposed RS3DAda method. "SA" indicates South America and "NA" indicates North America.}
  \label{fig:3d_vis}
\end{figure*}
\end{center}

\section{Introduction}
    \par In recent years, advancements in machine learning have significantly enhanced the interpretation of Earth observation (EO) data. A key task is 3D semantic reconstruction from 2D monocular remote sensing (RS) images, vital for environmental monitoring, urban planning, and disaster response \cite{li20193d, 3dbuilidng1, 3dbuilding2, 3dbuilding3}. This technique creates a 3D scene with semantic information from single-view images, unlike point cloud-based 3D reconstruction, which uses LiDAR or stereo cameras. Monocular 3D reconstruction is more scalable and requires less expensive equipment, making it suitable for global applications \cite{joint_height, joint_height4, jonint_height2, joint_height3,li20193d, 3dbuilidng1, 3dbuilding2, 3dbuilding3, building_height2, gordon2020learning, img2dsm, building_height}. This task combines land cover mapping, which is closely related to semantic segmentation in computer vision, and height estimation. However, acquiring RS annotations is costly and time-consuming, especially for high-resolution height data obtained through satellite LiDAR (e.g., GEDI, ICESat) \cite{sohn2004extraction,hermosilla2011evaluation,li2020high} or multi-view matching \cite{ameri2002high,zhang2003multi,yu2021automatic,liu2023deep,han2020state,mahphood2019dense}. Additionally, high-resolution land cover mapping datasets \cite{demir2018deepglobe,wang2021loveda,xia2023openearthmap} often lack corresponding height data. Moreover, the availability of RS data is regionally skewed, with developed regions having abundant data and developing regions lacking high-resolution datasets. Schmitt et al. \cite{schmitt2023there} reviewed over 380 RS datasets, revealing that few datasets come from Oceania, South America, Africa, and Asia, while most originate from Europe and North America. This geographic limitation in RS datasets raises a crucial question: Can findings from numerous research papers be applied to these underrepresented regions?

The aforementioned challenges can be effectively addressed using synthetic data. Current 3D modeling technology has the potential to create various landscape features with accompanying land cover semantic labels and height values. Therefore, we present \textit{SynRS3D}, a high-quality, high-resolution synthetic RS 3D dataset. SynRS3D includes 69,667 images with various ground sampling distances (GSD) ranging from 0.09 to 1 meter, and annotations for height estimation and land cover mapping across diverse scenes. However, models trained solely on synthetic data tend to overfit to these datasets, resulting in significantly reduced performance when applied to real-world environments due to the large domain gap. Existing synthetic datasets \cite{bourdis2011constrained,zou2020game,kong2020synthinel,shermeyer2021rareplanes,xu2022simpl,reyes2022syntcities,reyes20232d,song2024syntheworld,xiong2023benchmark} often exhibit this significant performance gap compared to models trained on real data.

To bridge this gap, we introduce \textit{RS3DAda}, a novel baseline aimed at advancing research on SynRS3D and setting a benchmark for multi-task UDA from RS synthetic to real scenarios. This approach utilizes a self-training framework and incorporates a land cover branch to enhance the quality of pseudo-labels of the height estimation branch, thus stabilizing RS3DAda training and boosting the accuracy of both branches. For the height estimation task, our model even outperforms models trained on real-world data in the challenging areas. Figure~\ref{fig:3d_vis} shows the results of 3D semantic reconstruction globally using models trained with our RS3DAda method on the SynRS3D dataset. Furthermore, we include disaster mapping outcomes for earthquake and wildfire scenarios using our model in Appendix~\ref{sec:disaster}, showcasing its efficacy in real-world disaster response applications.

The major contributions of this work can be summarized in three aspects. First, we propose SynRS3D, the largest RS synthetic dataset with comprehensive annotations and geographic diversity for RS tasks. Second, we design RS3DAda, a robust and effective multi-task UDA algorithm for land cover mapping and height estimation. Third, based on SynRS3D, we benchmark various scenarios for RS, including synthetic-only scenarios, combining SynRS3D with real data scenarios, and transferring SynRS3D to real scenarios for land cover mapping and height estimation. We hope our dataset and proposed method to advance the progress of synthetic learning in RS applications.

\section{Related Work}
\noindent \textbf{High-Resolution Earth Observation.} High-resolution remote sensing technology enables us to capture images with a GSD of less than 1 meter, significantly enhancing our understanding of Earth's surface details. Deep learning has become a powerful tool for analyzing these images. High-resolution imagery allows for precise land cover mapping \cite{li2023synergistical, zhou2023dynamic, hong2023cross, wang2023deep, liu2023rethinking, xiao2023novel}. Concurrently, height estimation research \cite{building_height2, gordon2020learning, img2dsm, building_height} focuses on determining accurate surface elevations. Some studies \cite{joint_height, joint_height4, jonint_height2, joint_height3} combine these tasks, training models for both land cover mapping and height estimation simultaneously. Most 3D reconstruction studies use real-world data, concentrating on buildings and often neglecting valuable classes like trees \cite{li20193d, 3dbuilidng1, 3dbuilding2, 3dbuilding3}. Multi-view RS for 3D reconstruction \cite{multiview3d_5, multiview3d_4, multiview3d_3, multiview3d_2, multiview3d_1, multiview3d_6} is expensive and geographically limited. Benchmark datasets \cite{demir2018deepglobe, wang2021loveda, xia2023openearthmap, dfc18, dfc2019, dfc2023, ogc, rottensteiner2012isprs} have been constructed for model training; however, acquiring high-resolution RS data is costly and time-consuming due to manual labeling and sophisticated equipment requirements. This limits the number of samples available for training robust models. Additionally, real datasets often lack geographic diversity, which can hinder model generalization to new, unseen areas.

\noindent \textbf{Synthetic Remote Sensing Dataset.} Traditional methods for creating synthetic data utilize deep learning generative models, such as diffusion models \cite{ho2020denoising} and generative adversarial networks (GANs) \cite{goodfellow2014generative}, alongside 3D modeling techniques. Generative models often struggle to produce data outside their training distribution and lack the precise control needed for RS tasks. In contrast, 3D modeling approaches in computer graphics, which leverage game engines or 3D software \cite{butler2012naturalistic, mayer2016large, ros2016synthia, richter2016playing, handa2016scenenet, wrenninge2018synscapes}, have shown more success. However, creating synthetic data for RS is inherently more complex. A single 1024x1024 RS image demands hundreds of buildings, thousands of trees, and diverse topological features like rivers and roads, unlike street views requiring fewer assets. Recent studies have employed 3D software for synthesizing RS data with automatic annotations \cite{bourdis2011constrained, zou2020game, kong2020synthinel, shermeyer2021rareplanes, xu2022simpl, reyes2022syntcities, reyes20232d, song2024syntheworld, xiong2023benchmark}. Despite their utility, these datasets often suffer from limited geographic diversity \cite{bourdis2011constrained,zou2020game,xu2022simpl,reyes2022syntcities,reyes20232d,xiong2023benchmark}, semantic categories \cite{bourdis2011constrained,zou2020game,kong2020synthinel,shermeyer2021rareplanes,xu2022simpl,reyes2022syntcities,reyes20232d,xiong2023benchmark}, and comprehensive semantic and height information \cite{bourdis2011constrained,zou2020game,kong2020synthinel,shermeyer2021rareplanes,xu2022simpl,song2024syntheworld}, impacting their effectiveness in training robust, globally applicable models.

\noindent \textbf{Unsupervised Domain Adaptation (UDA).} UDA aims to adapt a model trained on labeled data from a source domain to perform well on an unlabeled target domain, reducing the constraints and costs of data annotation. For semantic segmentation, most works focus on adversarial learning \cite{tsai2019domain, vu2019advent, hoffman2016fcns, hoffman2018cycada, gong2021dlow, wang2019transferable, wang2020classes, tsai2018learning, luo2019taking} and self-training \cite{zou2018unsupervised, lian2019constructing, zhang2019category, zou2019confidence, mei2020instance, sakaridis2018model, yang2020fda, Vu_2019_ICCV, hoyer2022daformer, hoyer2022hrda, hoyer2023mic}. Unlike semantic segmentation tasks, height estimation is a regression task, aligning closely with monocular depth estimation. For UDA in monocular depth estimation, methods often focus on image translation to reduce domain gaps \cite{atapour2018real, zhao2019geometry, zheng2018t2net, chen2019crdoco}, and some use self-training with pseudo-labels \cite{lopez2023desc, yang2021self, yen20223d}. In RS, most research \cite{iqbal2020weakly, lu2019multisource, othman2017domain, tasar2020standardgan, tong2020land, zhao2023semantic} focuses on applying developed techniques from computer vision to adapt models trained on real-world data to different real-world environments (real-to-real). However, only a small number of studies \cite{liu2024source,xie2024multimodal} investigate the challenges of adapting synthetic data to real-world environments (synthetic-to-real). To the best of our knowledge, RS3DAda is the first work to explore UDA algorithms specifically designed for synthetic-to-real domain adaptation for multi-task dense prediction in RS.

\section{The SynRS3D Acquisition Protocol and Statistical Analysis}
\begin{table}[t]
\caption{Comparisons of various RS synthetic datasets based on diversity, image capture details, asset origins, tasks, and the number of images. The diversity is categorized into City-Replica (datasets mimicking specific cities) and Style-Extended (covering a range of urban styles). Image capture attributes include GSD, resolution, and perspective (Nadir, Oblique). Asset origins are denoted as Manually-made (M), Game-origin (G), Procedurally-generated (P), and Real (R). Tasks cover Change Detection (CD), Building Segmentation (BS), Object Detection (OD), Disparity Estimation (DE), Height Estimation (HE), Land Cover (LC), and Building Change Detection (BCD). }
\label{tab:datasets_comparision}
\centering
\begin{adjustbox}{width=\textwidth,center}
\begin{tabular}{lcccccccccc}
\toprule
\multirow{2}{*}{RS Synthetic Datasets} & \multicolumn{2}{c}{Diversity} & \multicolumn{3}{c}{Image Capture} & \multicolumn{3}{c}{Assets} & \multirow{2}{*}{Task} & \multirow{2}{*}{\# Images} \\
\cmidrule(lr){2-3} \cmidrule(lr){4-6} \cmidrule(lr){7-9}
 & City-Replica & Style-Extended & GSD (m) & Image Size & Perspective & Layout & Geometry & Texture & & \\
\midrule
AICD \cite{bourdis2011constrained}               & \cmark & \xmark           & $\ast$      & $800\times600$         & Nad., Obl.  & M & M & M & CD & $\sim1K$ \\
GTA-V-SID \cite{zou2020game}          & \cmark & \xmark           & $1.0$ & $500\times500$ & Nad.       & G & G & G & BS & $\sim0.12K$  \\
Synthinel-1 \cite{kong2020synthinel}      & \cmark & \cmark           & $0.3$ & $572\times572$ & Nad.       & R & M & M+R & BS & $\sim1K$  \\
RarePlanes \cite{shermeyer2021rareplanes}         & \cmark & \xmark           & $0.31\sim0.39$ & $512\times512$ & Nad., Obl. & R & M & M+R & OD & $\sim65K$  \\
SyntCities \cite{reyes2022syntcities}        & \cmark & \xmark & $0.1, 0.3, 1.0$      & $1024\times1024$         & Nad.        & R & M & M+R & DE & $\sim8K$ \\
GTAH  \cite{xiong2023benchmark}             & \cmark & \xmark           & $\ast$    & $1920\times1080$ & Nad., Obl.  & G & G & G & HE & $\sim 28.6K$  \\
SyntheWorld \cite{song2024syntheworld}        & \xmark           & \cmark & $0.3\sim1.0$ & Various & Nad., Obl. & P & P+M & P & LC, BCD & $\sim40K$  \\
SMARS \cite{reyes20232d}             & \cmark            & \xmark           & $0.3, 0.5$      &   Various       & Nad.        & R & M & M+R & LC, HE, BCD                   & $4$     \\
\midrule
\rowcolor{blue!20}SynRS3D (Ours) & \xmark & \cmark & $0.09\sim1$ & $512\times512$   & Nad., Obl.  & P & P+M & P & LC, HE, BCD   & $\sim70K$   \\
\bottomrule
\end{tabular}
\end{adjustbox}
\end{table}

Although simulating remote sensing scenes poses significant challenges due to the need for numerous assets, we mitigate this issue using procedural modeling techniques \cite{musgrave1989synthesis,muller2006procedural,kim2018procedural}. Instead of manually modeling each element, we incorporate rules derived from real-world knowledge, formalized into scripts. The generation system is controlled via hyperparameters like city style, asset ratio, and texture style, allowing us to create a high-quality, diverse synthetic dataset. Table~\ref{tab:datasets_comparision} compares existing synthetic RS datasets with SynRS3D, highlighting our dataset's advantages in diversity, functionality, and scale.

\subsection{Statistics for SynRS3D}

\begin{figure*}[t!]
  \centering
  \includegraphics[width=\textwidth]{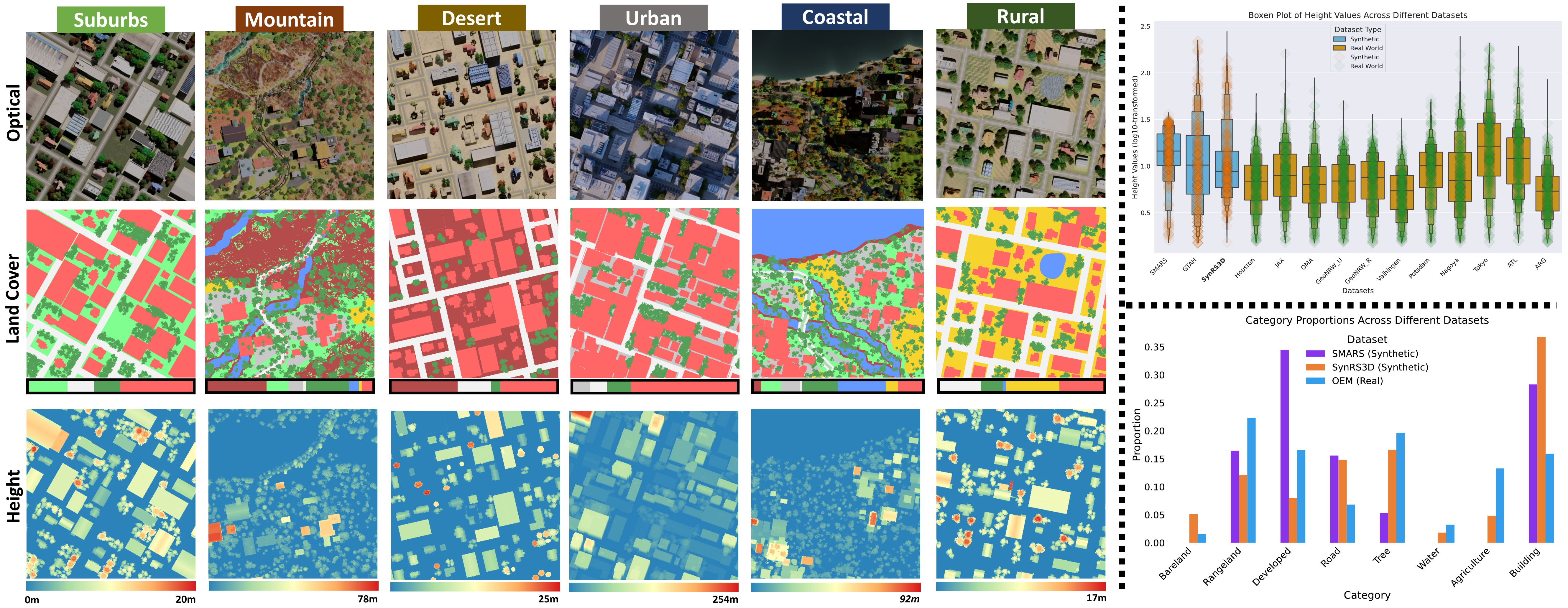}
  \caption{Examples and statistics of SynRS3D.}
  \label{fig:statistical}
\end{figure*}

The left section of Figure~\ref{fig:statistical} showcases RGB images, land cover labels, and height maps for six styles in SynRS3D, representing diverse real-world environments. The upper right section compares the height distribution of SynRS3D with two leading synthetic datasets, SMARS \cite{reyes20232d} and GTAH \cite{xiong2023benchmark}, as well as 11 real-world height datasets. SynRS3D's height distribution closely matches real-world data, while SMARS and GTAH show limitations. Specifically, SMARS, which mimics Paris and Venice, has a narrow height range. GTAH, based on the game GTAV, which mimics Los Angeles, shows a wider height range but with larger mean and variance, making it less representative of other cities. SynRS3D was constructed using the following prior knowledge \cite{worldbank2024}: backward regions (low buildings) cover about 12\% of the world's areas, emerging regions (mid buildings) cover about 70\%, and developed regions (tall buildings) cover about 18\%. The bottom right section contrasts the land cover proportions in SynRS3D with those in SMARS and the real-world OEM dataset \cite{xia2023openearthmap}, which is currently the largest and most geographically diverse dataset. SynRS3D's land cover categories—including Bareland, Rangeland, Developed Space, Roads, Trees, Water, Agricultural Land, and Buildings—match well with the OEM dataset. In contrast, SMARS has only five categories, limiting its effectiveness for comprehensive land cover mapping.

\subsection{Generation Workflow of SynRS3D}
\label{sec:workflow}
\begin{center}
\begin{figure*}[t]
  \centering
  \includegraphics[width=\textwidth]{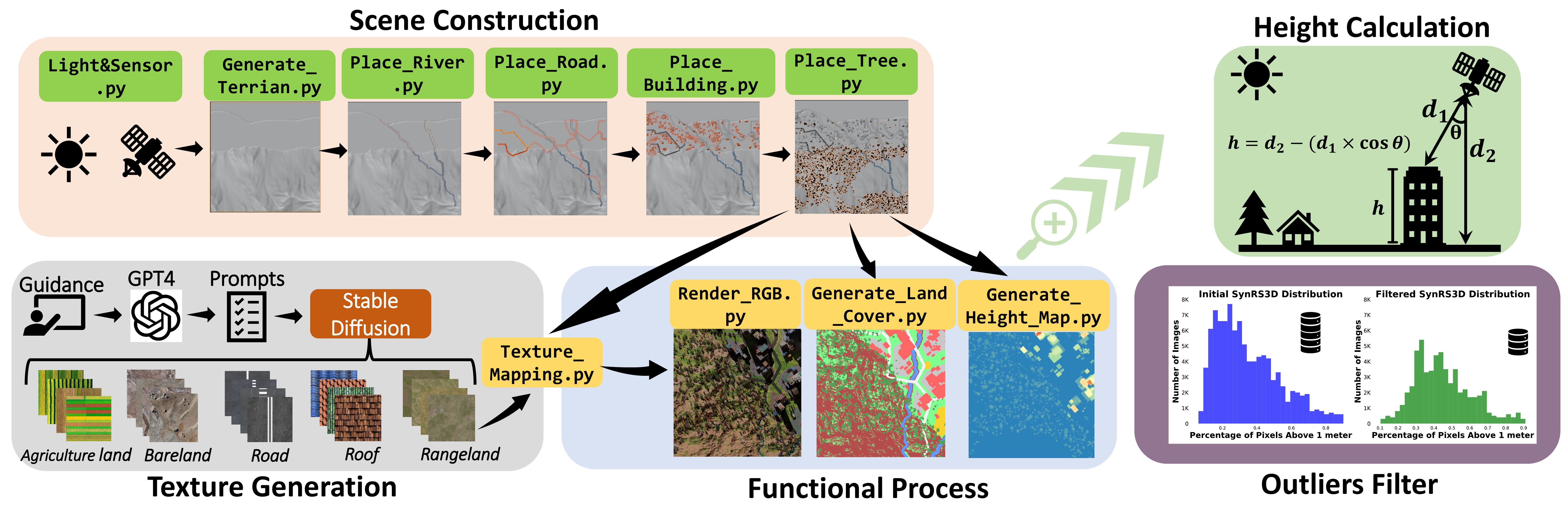}
  \caption{Generation workflow of SynRS3D.}
  \label{fig:generation_workflow}
\end{figure*}
\end{center}

The generation process of SynRS3D employs tools such as Blender \cite{blender}, Python, GPT-4 \cite{openai2023gpt4}, and Stable Diffusion \cite{rombach2022high}, as illustrated in Figure~\ref{fig:generation_workflow}. It begins with Python scripts that translate synthetic scene rules into parameter-controlled instructions for tasks like terrain generation, sensor placement, and asset placement. The geometry of the buildings and trees is created both procedurally and manually.
Stable Diffusion generates textures based on detailed text prompts from GPT-4, ensuring high-quality and diverse textures. Blender's compositor node and Python scripts then generate accurate land cover labels and height maps. The specifics of the height map generation are detailed in the height calculation section of Figure~\ref{fig:generation_workflow}. Our dataset's height maps are produced within the 3D software using simple geometric algorithms, resulting in a completely accurate normalized Digital Surface Model (nDSM). An nDSM represents the height of objects above the ground, providing clear information about buildings and vegetation. In contrast, the height maps for the comparative datasets GTAH and SMARS are Digital Surface Models (DSM), which include the ground and objects. Converting DSM to nDSM requires additional processing using the dsm2dtm~\footnote{https://github.com/seedlit/dsm2dtm} algorithm, which introduces noises. Optical images are produced using rendering scripts. To generate building change detection masks, we follow a structured process. Initially, buildings are randomly removed from scenes, and textures are reapplied to create pre-event images. Subsequently, land cover labels are subtracted to produce the change detection masks. After the initial dataset generation, images with anomalous height distributions are filtered out. This step ensures that the final version of SynRS3D closely aligns with real-world height distributions. The specific filtering algorithm used and examples of building change detection masks can be found in Appendix~\ref{sec:workflow}.

\vspace{-10mm}
\section{Multi-Task Domain Adaptation for Remote Sensing Tasks (RS3DAda)}
SynRS3D features low costs, high diversity, and large volume. However, there is a clear domain gap between synthetic data and real-world environments, limiting the use of SynRS3D. This limitation is particularly evident in RS, where synthetic-to-real UDA algorithms are lacking. To bridge this gap, we developed RS3DAda, which leverages land cover labels and height values to complement each other. In addition, it harnesses the potential of unlabeled real-world data, establishing the first benchmark for synthetic-to-real RS-specific multi-task UDA.

\noindent \textbf{Basic Framework.} In this work, we adopt self-training~\cite{lee2013pseudo} as our basic UDA technique due to its superior stability and adaptability across both land cover mapping and height estimation tasks. Additionally, the teacher-student framework~\cite{tarvainen2017mean} is incorporated into the method to enhance performance and generalizability by stabilizing training with the exponential moving average of model weights. The synthetic dataset image set \(\mathcal{X}_s\) serves as the source domain, with access to corresponding land cover labels \(\mathcal{Y}_{LC}^s\) and height maps \(\mathcal{Y}_{H}^s\). The real dataset image set \(\mathcal{X}_t\) serves as the target domain without any access to the labels.

\begin{figure*}[t!]
  \centering
  \includegraphics[width=\textwidth]{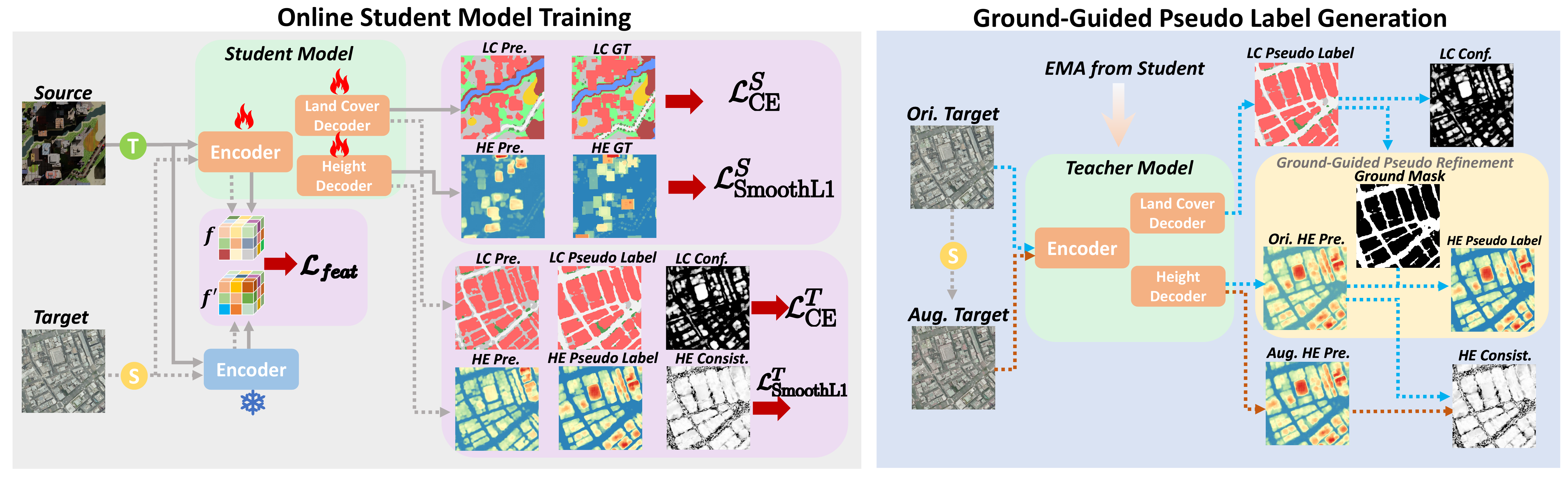}
  \caption{Overview of the proposed RS3DAda method. \textcolor{T}{T} denotes statistical image translation, \textcolor{S}{S} represents strong augmentation. For Online Student Model Training, \textcolor{linegray}{dotted line:} target image, \textcolor{linegray}{solid line:} source image. For Ground-Guided Pesudo Label Generation, \textcolor{lineblue}{dotted line:} original target image, \textcolor{linered}{dotted line:} strong augmented target image.}
  \label{fig:mtda}
\end{figure*}

\noindent \textbf{Source Domain Training.} In this stage, shown in the left part of Figure~\ref{fig:mtda}, source domain images \(\mathcal{X}_s\) undergo statistical image translation using simple Histogram Matching (HM) and Pixel Distribution Matching (PDA), which follow the conclusion of \cite{abramov2020keep}. This process aligns the styles of source images with the target domain, resulting in translated images \(\mathcal{X}_s'\). These translated images from the source domain are then fed into the student network, producing predicted land cover labels \(\hat{Y}_{LC}^s\) and height \(\hat{Y}_{H}^s\). The supervised loss for the source domain is defined as:
\begin{equation}
    \mathcal{L}_{source} = \frac{1}{N} \sum_{i=1}^{N} \left(\mathcal{L}_{CE}(\hat{Y}_{LC}^{s (i)}, \mathcal{Y}_{LC}^{s (i)}) + \mathcal{L}_{SmoothL1}(\hat{Y}_{H}^{s (i)}, \mathcal{Y}_{H}^{s (i)})\right),
\end{equation}
where \(\mathcal{L}_{CE}\) is the cross-entropy loss, \(\mathcal{L}_{SmoothL1}\) is the Smooth L1 loss, and \(N\) is the total number of samples.

\noindent \textbf{Land Cover Pseudo-Label Generation.} To generate high-quality pseudo-labels for land cover mapping, as illustrated in the right section of Figure~\ref{fig:mtda}, target domain images \(\mathcal{X}_t\) are strongly augmented denoted as \(\mathcal{X}_t'\). We adopt color jitter, Gaussian blur, and ClassMix \cite{olsson2021classmix} as the strong augmentations. The teacher model then predicts the land cover pseudo-labels \(\tilde{Y}_{LC}^t\). After that, we use a threshold \(\tau\) to generate a confidence map \(\mathcal{C}_{LC}= \mathbb{I}(\tilde{Y}_{LC}^t > \tau)\), where \(\mathbb{I}\) is the indicator function.

\noindent \textbf{Height Pseudo-Label Generation.}
\label{sec:methods}
Height pseudo-labels are generated by leveraging prior knowledge that only trees and buildings have height values. This is the first attempt to use ground information to correct height pseudo-labels, inspired by the empirical observations that the network achieves superior accuracy on the ground class early in the training stage. We refine height pseudo-labels using a ground mask \(\mathcal{G}\) generated from the land cover mapping branch as \(\mathcal{G} = \mathbb{I}(\tilde{Y}_{LC}^t = \text{Ground})\), and refine the height pseudo-labels by \(\tilde{Y}_{H}^{t, refined} = \tilde{Y}_{H}^{t, ori} \cdot (1 - \mathcal{G})\). We also create a height consistency map \(\mathcal{C}_{H}\):
\begin{equation}
    \mathcal{C}_{H} = \mathbb{I}\left(\max\left(\frac{\tilde{Y}_{H}^{t, ori}}{\tilde{Y}_{H}^{t, aug}}, \frac{\tilde{Y}_{H}^{t, aug}}{\tilde{Y}_{H}^{t, ori}}\right) \leq \eta\right),
\end{equation}
indicating that we consider predictions reliable if they remain stable under perturbations, where \(\eta\) is the threshold.

\noindent \textbf{Target Domain Training.} The target domain training loss is:
\begin{equation}
    \mathcal{L}_{target} = \frac{1}{N} \sum_{i=1}^{N} \left(\mathcal{L}_{CE}(\hat{Y}_{LC}^{t (i)}, \tilde{Y}_{LC}^{t (i)}) \cdot \mathcal{C}_{LC}^{(i)} + \mathcal{L}_{SmoothL1}(\hat{Y}_{H}^{t (i)}, \tilde{Y}_{H}^{t, refined (i)}) \cdot \mathcal{C}_{H}^{(i)}\right),
\end{equation}

\noindent \textbf{Feature Constraint.} To alleviate overfitting, we adopt a frozen DINOv2~\cite{oquab2023dinov2} encoder to supervise the student encoder's updates, inspired by \cite{yang2024depth}. Let \(\mathbf{f}\) denote the features of the student encoder and \(\mathbf{f}'\) denote the features of the frozen DINOv2 encoder. We utilize cosine similarity to constrain the feature updates with the loss defined as:
\begin{equation}
    \mathcal{L}_{feat} = \begin{cases} 
    1 - \frac{\mathbf{f} \cdot \mathbf{f}'}{\|\mathbf{f}\| \|\mathbf{f}'\|} & \text{if } \frac{\mathbf{f} \cdot \mathbf{f}'}{\|\mathbf{f}\| \|\mathbf{f}'\|} < \epsilon \\
    0 & \text{otherwise}
    \end{cases}
\end{equation}
where \(\epsilon\) is the threshold.

\noindent \textbf{Overall Loss.} The overall loss of the model is \(\mathcal{L}_{overall} = \mathcal{L}_{source} + \lambda_{target} \mathcal{L}_{target} + \lambda_{feat} \mathcal{L}_{feat}\), where \(\lambda_{target}\) and \(\lambda_{feat}\) are weighting coefficients for the loss terms.

\section{Experiments}
\noindent \textbf{Evaluation Datasets.} We evaluate our synthetic dataset, SynRS3D, from various aspects. Table~\ref{table:realworld_datasets} (a) shows the real-world height estimation datasets, while Table~\ref{table:realworld_datasets} (b) lists the real-world land cover mapping datasets. In Section~\ref{sec:souceonly}, we compare SynRS3D with other synthetic datasets under the source-only setting, demonstrating its smaller domain gap and potential for direct usage in the real world. In Section~\ref{sec:synrs3dandreal}, we investigate the advantages of SynRS3D for augmenting real-world data through fine-tuning and joint training. Specifically, in Section~\ref{sec:expRS3DAda}, to evaluate the effectiveness of RS3DAda, we divide the 11 height estimation datasets into two target domains: \textit{Target Domain 1} includes 6 widely-used public datasets, while \textit{Target Domain 2} contains 5 more challenging datasets to assess and contrast the generalization ability of SynRS3D with real datasets. 
\begin{table}[ht]
\caption{Datasets setup for the experiments on height estimation and land cover mapping. (a) \textbf{Height Estimation Datasets}: We detail 11 height estimation datasets categorized into two target domains. The first six datasets for training the model with real-world data are sourced from Europe and the United States as shown in Section~\ref{sec:expRS3DAda}. The remaining five datasets covering more challenging areas are characterized by: {\color{orange}{notable height mean\&standard deviation}}, {\color{cyan}{non-RGB channels}}, and {\color{purple}{varied regions}} outside of the US and EU, used for evaluation in Section~\ref{sec:expRS3DAda}. (b) \textbf{Land Conver Mapping Datasets}: We evaluate our method on the five commonly used datasets covering diverse environments.}
\label{table:realworld_datasets}
\centering
\begin{adjustbox}{width=\textwidth,center}
\begin{tabular}{cc}
\begin{subtable}[t]{0.9\textwidth}
\centering
\begin{tabular}{c|c|c|c|c}
\toprule
\multicolumn{5}{c}{\textbf{Real-World Height Estimation Datasets}} \\
\toprule
\textbf{Types} & \textbf{Datasets} & \textbf{Region} & \textbf{Height mean\&std} & \textbf{Channel} \\
\midrule
\multirow{6}{*}{\begin{tabular}[c]{@{}c@{}}Target \\ Domain 1\end{tabular}} 
 & Houston \cite{dfc18} & US & \texttt{[}3.07, 5.02\texttt{]} & RGB \\
 & JAX \cite{dfc2019} & US & \texttt{[}4.73, 9.02\texttt{]} & RGB \\
 & OMA \cite{dfc2019} & US & \texttt{[}2.37, 5.27\texttt{]} & RGB \\
 & GeoNRW\_Urban \cite{baier2021synthesizing}& Germany & \texttt{[}2.46, 4.31\texttt{]} & RGB \\
 & GeoNRW\_Rural \cite{baier2021synthesizing}& Germany & \texttt{[}2.03, 4.21\texttt{]} & RGB \\
 & Potsdam \cite{rottensteiner2012isprs} & Germany & \texttt{[}3.02, 5.68\texttt{]} & RGB \\
\midrule
\multirow{5}{*}{\begin{tabular}[c]{@{}c@{}}Target \\ Domain 2\end{tabular}} 
 & ATL \cite{ogc}& US & \color{orange}{\texttt{[}8.40, 13.41\texttt{]}} & RGB \\
 & ARG \cite{ogc}& \textcolor{purple}{Argentina} & \texttt{[}3.90, 4.29\texttt{]} & RGB \\
 & Nagoya \cite{aw3d2018} & \textcolor{purple}{Japan} & \color{orange}{\texttt{[}7.36, 11.84\texttt{]}} & RGB \\
 & Tokyo \cite{aw3d2018} & \textcolor{purple}{Japan} & \color{orange}{\texttt{[}15.73, 22.77\texttt{]}} & RGB \\
 & Vaihingen \cite{rottensteiner2012isprs} & Germany & \texttt{[}2.36, 3.57\texttt{]} & \textcolor{cyan}{NIR, G, B} \\
\bottomrule
\end{tabular}
\caption{Real-world height estimation datasets.}
\end{subtable}
&
\begin{subtable}[t]{0.8\textwidth}
\centering
\begin{tabular}{c|c|c|c}
\toprule
\multicolumn{4}{c}{\textbf{Real-World Land Cover Mapping Datasets}} \\
\toprule
\textbf{Types} & \textbf{Datasets} & \textbf{Region} & \textbf{Categories} \\
\midrule
\multirow{5}{*}{\begin{tabular}[c]{@{}c@{}}Target \\ Domain\end{tabular}} 
 & OEM \cite{xia2023openearthmap}& Global & 8 \\
 & Vaihingen \cite{rottensteiner2012isprs}& Germany & 6 \\
 & Potsdam \cite{rottensteiner2012isprs}& Germany & 6 \\
 & JAX \cite{dfc2019}& US & 6 \\
 & OMA \cite{dfc2019}& US & 6 \\
\bottomrule
\end{tabular}
\caption{Real-world land cover mapping datasets.}
\end{subtable}
\end{tabular}
\end{adjustbox}
\end{table}

\noindent \textbf{Evaluation Metrics.} We employ Intersection over Union (IoU) and Mean Intersection over Union (mIoU) as the metrics to evaluate the model's performance for land cover mapping tasks. For height estimation, we use Mean Absolute Error (MAE), Root Mean Squared Error (RMSE), and accuracy metrics \cite{eigen2014depth} denoted as \(\delta\), along with our custom metric \(F_{1}^{HE}\). Detailed metric definitions are in Appendix~\ref{sec:metrics}.

\noindent \textbf{Implementation Details.} Unless specifically detailed, all experiments utilize the pre-trained DINOv2~\cite{oquab2023dinov2} implemented with ViT-L~\cite{dosovitskiy2020image} as the encoder, with DPT \cite{ranftl2021vision} serving as both the land cover mapping and height estimation decoders. The hyperparameters for the various experiments are provided in Appendix~\ref{sec:setting}.

\subsection{Source-only Scenarios}
\begin{figure}[H]
  \centering
  \begin{minipage}{0.5\textwidth}
    \includegraphics[width=\textwidth]{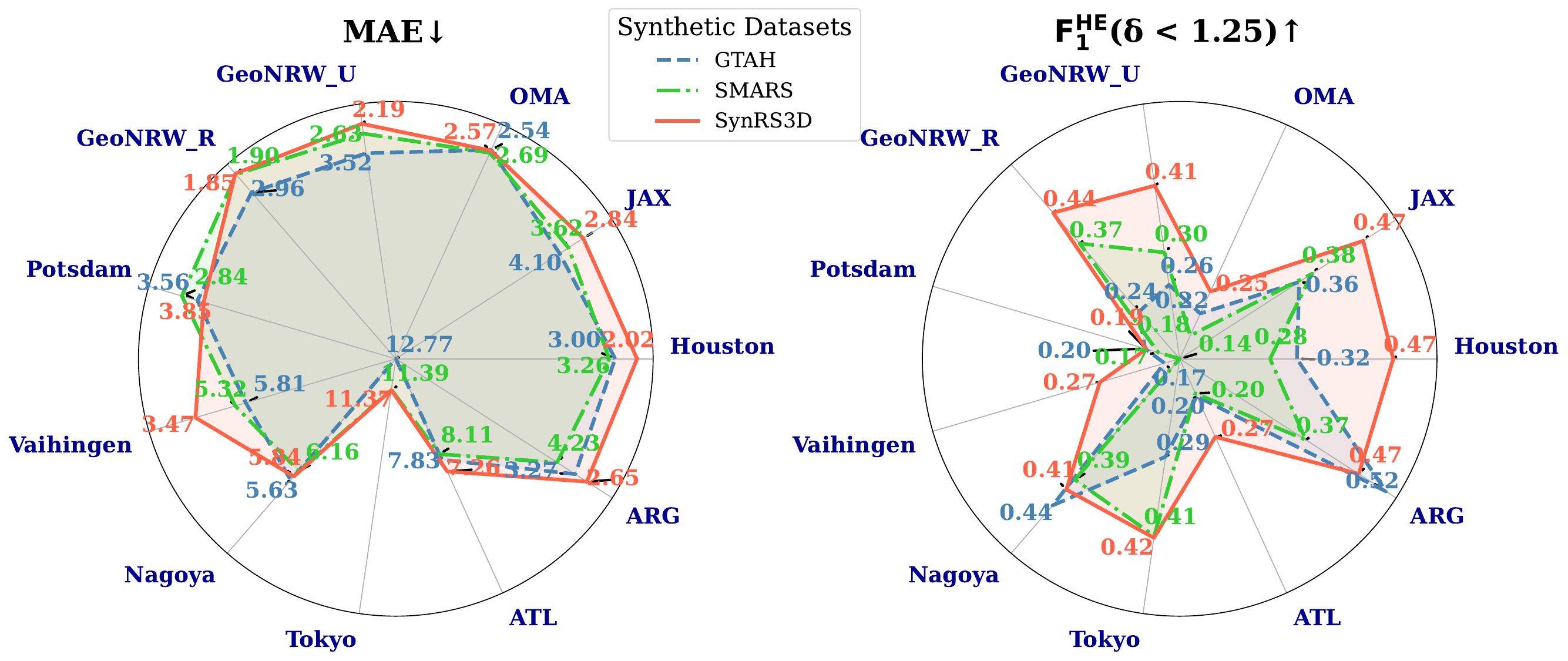}
    \caption{Source-only height estimation comparison of SynRS3D and other synthetic datasets showing in different metrics.}
    \label{fig:zero_shot_he_radar}
  \end{minipage}\hfill
  \begin{minipage}{0.47\textwidth}
    \noindent \textbf{Height Estimation.} We compared SynRS3D with other synthetic datasets in a source-only height estimation scenario. As shown in Figure~\ref{fig:zero_shot_he_radar}, SynRS3D outperformed competitors SMARS \cite{reyes20232d} and GTAH \cite{xiong2023benchmark} in 9 out of 11 real datasets. This superiority is attributed to the smaller domain gap and diversity of SynRS3D, as well as the precise calculation of heights within 3D software. In contrast, heights in SMARS and GTAH are not normalized, requiring additional algorithms for normalization, which introduces inherent noise in their height values.
  \end{minipage}
\end{figure}
\noindent \textbf{Land Cover Mapping.} We demonstrate the source-only capability of models trained on SynRS3D in the land cover mapping task. Table~\ref{tab:zero_shot_lc} compares SynRS3D with existing synthetic datasets, SMARS \cite{reyes20232d} and SyntheWorld \cite{song2024syntheworld}. Due to category inconsistencies, we present IoU and mIoU for shared categories including trees, buildings, and ground on the JAX \cite{dfc2019}, OMA \cite{dfc2019}, Vaihingen \cite{rottensteiner2012isprs}, and Potsdam \cite{rottensteiner2012isprs} datasets. The model trained on SynRS3D achieves the best results across these four real datasets using only land cover labels, examining the extraordinary compatibility of SynRS3D.
\begin{table}[h]
\caption{Source-only land cover mapping performance on various real-world datasets.}
\label{tab:zero_shot_lc}
\centering
\begin{adjustbox}{width=\textwidth, center}
\begin{tabular}{@{}l|cccc|cccc|cccc|cccc@{}}
\toprule
\multirow{2}{*}{Datasets} &
  \multicolumn{4}{c}{JAX \cite{dfc2019}} &
  \multicolumn{4}{c}{OMA \cite{dfc2019}} &
  \multicolumn{4}{c}{Vaihingen \cite{rottensteiner2012isprs}} &
  \multicolumn{4}{c}{Potsdam \cite{rottensteiner2012isprs} } \\ 
\cmidrule(lr){2-5} \cmidrule(lr){6-9} \cmidrule(lr){10-13} \cmidrule(lr){14-17}
 & Ground & Tree & Building & mIoU & Ground & Tree & Building & mIoU & Ground & Tree & Building & mIoU & Ground & Tree & Building & mIoU \\ 
\midrule
SMARS \cite{reyes20232d} &
  76.02 &
  43.13 &
  61.28 &
  60.14 &
  82.17 &
  17.25 &
  59.94 &
  53.12 &
  74.10 &
  58.40 &
  74.35 &
  68.95 &
  68.56 &
  5.35 &
  57.51 &
  43.81 \\
SyntheWorld \cite{song2024syntheworld} &
  74.63 &
  54.74 &
  64.18 &
  64.52 &
  81.29 &
  \textbf{45.83} &
  56.56 &
  61.23 &
  72.69 &
  68.09 &
  75.67 &
  72.15 &
  69.09 &
  32.49 &
  55.88 &
  52.49 \\
\textbf{SynRS3D} &
  \textbf{77.69} &
  \textbf{57.03} &
  \textbf{68.96} &
  \textbf{67.89} &
  \textbf{83.96} &
  41.08 &
  \textbf{62.28} &
  \textbf{62.44} &
  \textbf{75.66} &
  \textbf{68.58} &
  \textbf{79.61} &
  \textbf{74.61} &
  \textbf{74.26} &
  \textbf{35.34} &
  \textbf{69.46} &
  \textbf{59.69} \\
\bottomrule
\end{tabular}
\end{adjustbox}
\end{table}
\label{sec:souceonly}

\subsection{Combining SynRS3D with Real Data Scenarios}
\label{sec:synrs3dandreal}
\begin{figure}[t!]
\centering
\begin{subfigure}[t]{0.48\textwidth}
    \centering
    \includegraphics[width=\textwidth]{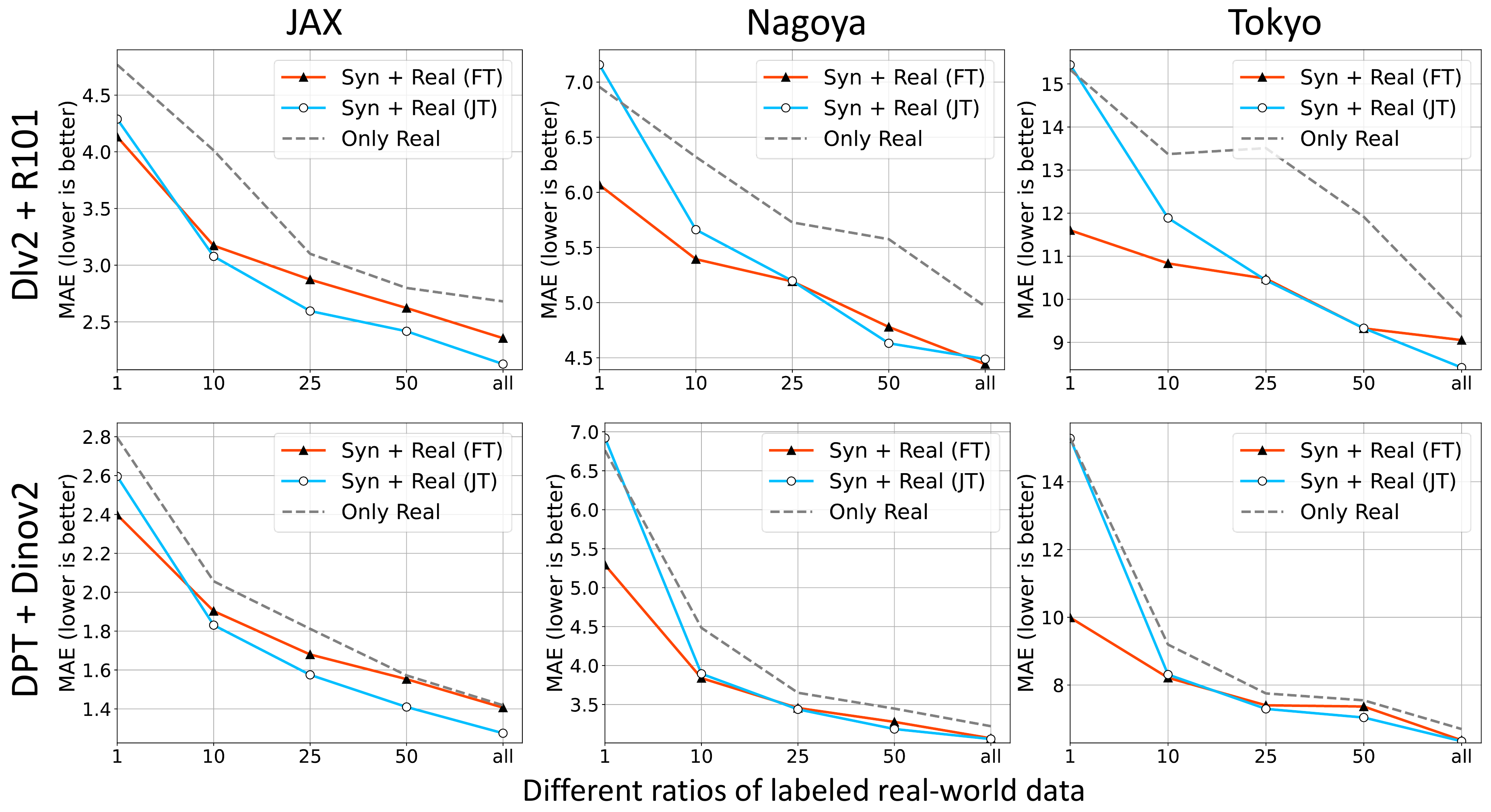}
    \caption{Height Estimation.}
    \label{fig:height_estimation}
\end{subfigure}
\hfill
\begin{subfigure}[t]{0.48\textwidth}
    \centering
    \includegraphics[width=\textwidth]{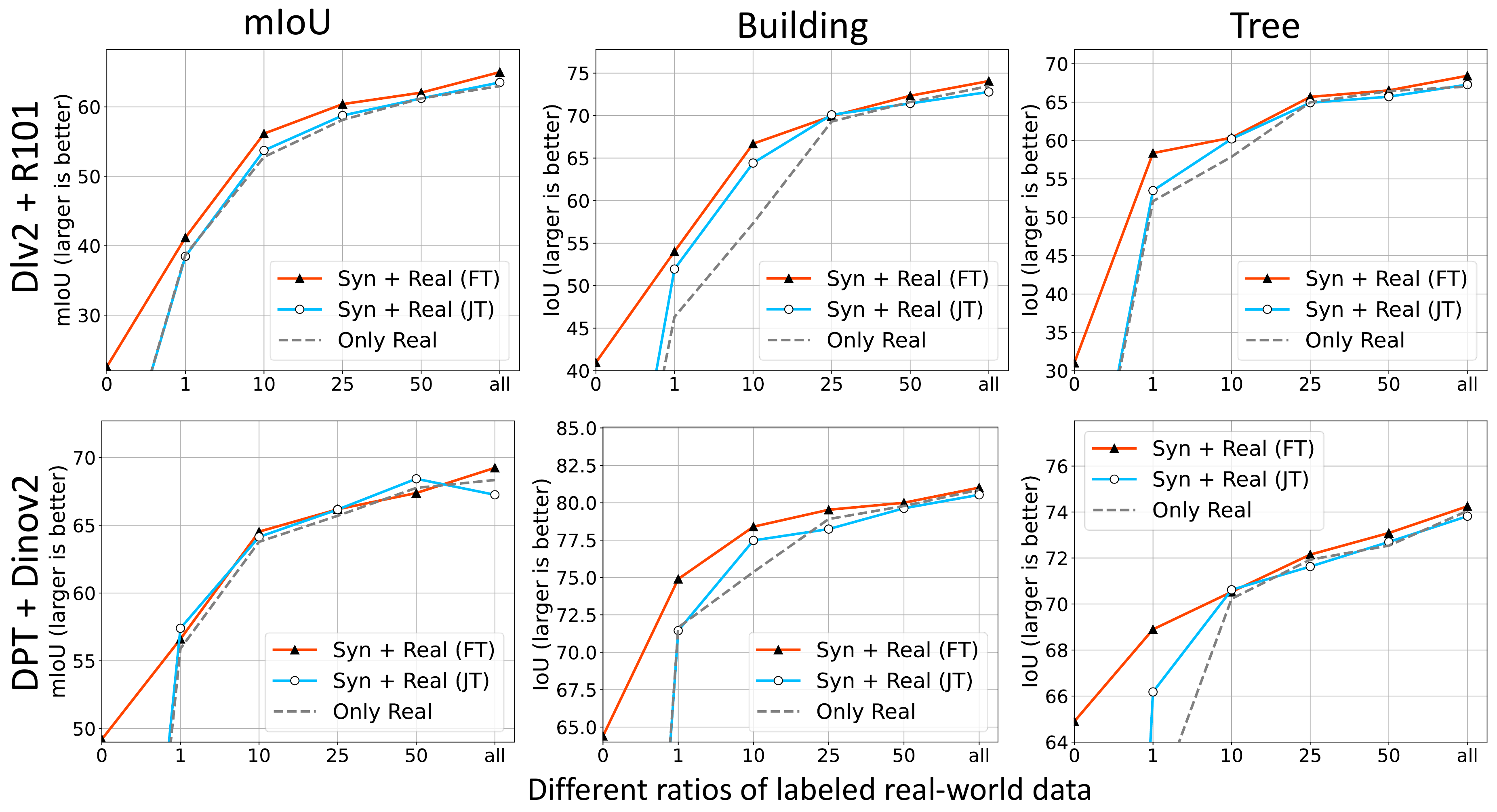}
    \caption{Land Cover Mapping.}
    \label{fig:land_cover_mapping}
\end{subfigure}
\caption{Performance evaluation of combining SynRS3D with real data in (a) height estimation and (b) land cover mapping. Height estimation is evaluated on various real datasets, and land cover mapping is evaluated on OEM dataset, showing IoU for building, tree, and mIoU. FT: fine-tuning on real data after pre-training on SynRS3D, JT: joint training with SynRS3D and real data.}
\label{fig:height_and_lcm}
\end{figure}

An important use of synthetic data is to augment real-world data. To demonstrate this capability of SynRS3D, we conducted two experiments. First, we trained models on SynRS3D and fine-tuned them on real data. Second, we combined SynRS3D with real data for joint training. We experimented with two different backbones: DINOv2 \cite{oquab2023dinov2}+DPT \cite{ranftl2021vision} and DeepLabV2 \cite{chen2017deeplab}+ResNet101 \cite{he2016deep}. Figure~\ref{fig:height_and_lcm} (a) showcases SynRS3D's performance in the height estimation task on three city datasets: JAX \cite{dfc2019}, Nagoya \cite{aw3d2018}, and Tokyo \cite{aw3d2018}. The results indicate that both approaches yield significant improvements when real data is scarce, with benefits diminishing as more real data is added. Additionally, the stronger the backbone, the smaller the improvement provided by SynRS3D, and vice versa. Figure~\ref{fig:height_and_lcm} (b) illustrates SynRS3D's augmentation capability in the land cover mapping task on OEM \cite{xia2023openearthmap} dataset, showing similar conclusions to the height estimation task.

\subsection{Transfer SynRS3D to Real-World Scenarios}
\label{sec:expRS3DAda}
\begin{figure}[h]
  \centering
  \begin{minipage}[h]{0.6\linewidth} 
    \centering
    \captionof{table}{Results of RS3DAda height estimation branch using DINOv2 \cite{oquab2023dinov2} and DPT \cite{ranftl2021vision}. The experimental results are divided as follows: 'Whole' denotes the evaluation results for the entire image. 'High' signifies the experimental results for image regions above 3 meters. T.D.1 and T.D.2 correspond to \textit{Target Domain 1} and \textit{Target Domain 2}, respectively, as specified in Table~\ref{table:realworld_datasets}. Avg. stands for the average value.}
    \label{tab:mtda_overall}
    \begin{adjustbox}{width=\textwidth,center}
      \begin{tabular}{l|cc|cc|ccc|ccc}
      \toprule
      {Model} & \multicolumn{2}{c}{{MAE $\downarrow$}} & \multicolumn{2}{c}{{RMSE $\downarrow$}} & \multicolumn{3}{c}{{Accuracy Metrics \cite{eigen2014depth} $\uparrow$}} & \multicolumn{3}{c}{{\(F_{1}^{HE}\) $\uparrow$}} \\
      \cmidrule(lr){2-3} \cmidrule(lr){4-5} \cmidrule(lr){6-8} \cmidrule(lr){9-11}
      \rowcolor{cyan!20}{\textcolor{gray!80}{\textbf{Avg. T.D.1}}} & \textcolor{gray!80}{\textbf{Whole}} & \textcolor{gray!80}{\textbf{High}} & \textcolor{gray!80}{\textbf{Whole}} & \textcolor{gray!80}{\textbf{High}} & \(\delta < 1.25\) & \(\delta < 1.25^2\) & \(\delta < 1.25^3\) & \(\delta < 1.25\) & \(\delta < 1.25^2\) & \(\delta < 1.25^3\) \\
      \midrule
      {Train-on-T.D.1} & \textbf{1.272} & \textbf{3.363} & \textbf{2.381} & \textbf{4.329} & \textbf{0.379} & \textbf{0.463} & \textbf{0.510} & \textbf{0.617} & \textbf{0.710} & \textbf{0.742} \\
      {Source Only} & 2.557 & 5.617 & 4.128 & 6.705 & 0.123 & 0.192 & 0.246 & 0.372 & 0.491 & 0.552 \\
      \textbf{RS3DAda} & 2.148 & 4.921 & 3.593 & 6.024 & 0.185 & 0.258 & 0.318 & 0.418 & 0.554 & 0.623 \\
      \midrule
      \rowcolor{orange!20}{\textcolor{gray!80}{\textbf{Avg. T.D.2}}} & \textcolor{gray!80}{\textbf{Whole}} & \textcolor{gray!80}{\textbf{High}} & \textcolor{gray!80}{\textbf{Whole}} & \textcolor{gray!80}{\textbf{High}} & \(\delta < 1.25\) & \(\delta < 1.25^2\) & \(\delta < 1.25^3\) & \(\delta < 1.25\) & \(\delta < 1.25^2\) & \(\delta < 1.25^3\) \\
      \midrule
      {Train-on-T.D.1} & 5.378 & 8.302 & 8.301 & 10.714 & 0.146 & 0.244 & 0.336 & 0.384 & 0.535 & 0.627 \\
      {Source Only} & 6.117 & 8.923 & 9.221 & 11.443 & 0.125 & 0.223 & 0.312 & 0.365 & 0.514 & 0.601 \\
      \textbf{RS3DAda} & \textbf{4.866} & \textbf{7.227} & \textbf{7.584} & \textbf{9.594} & \textbf{0.182} & \textbf{0.299} & \textbf{0.389} & \textbf{0.485} & \textbf{0.621} & \textbf{0.689} \\
      \bottomrule
      \end{tabular}
    \end{adjustbox}
  \end{minipage}
  \hspace{0.03\linewidth}
  \begin{minipage}[h]{0.35\linewidth} 
    \centering
    \includegraphics[width=\linewidth]{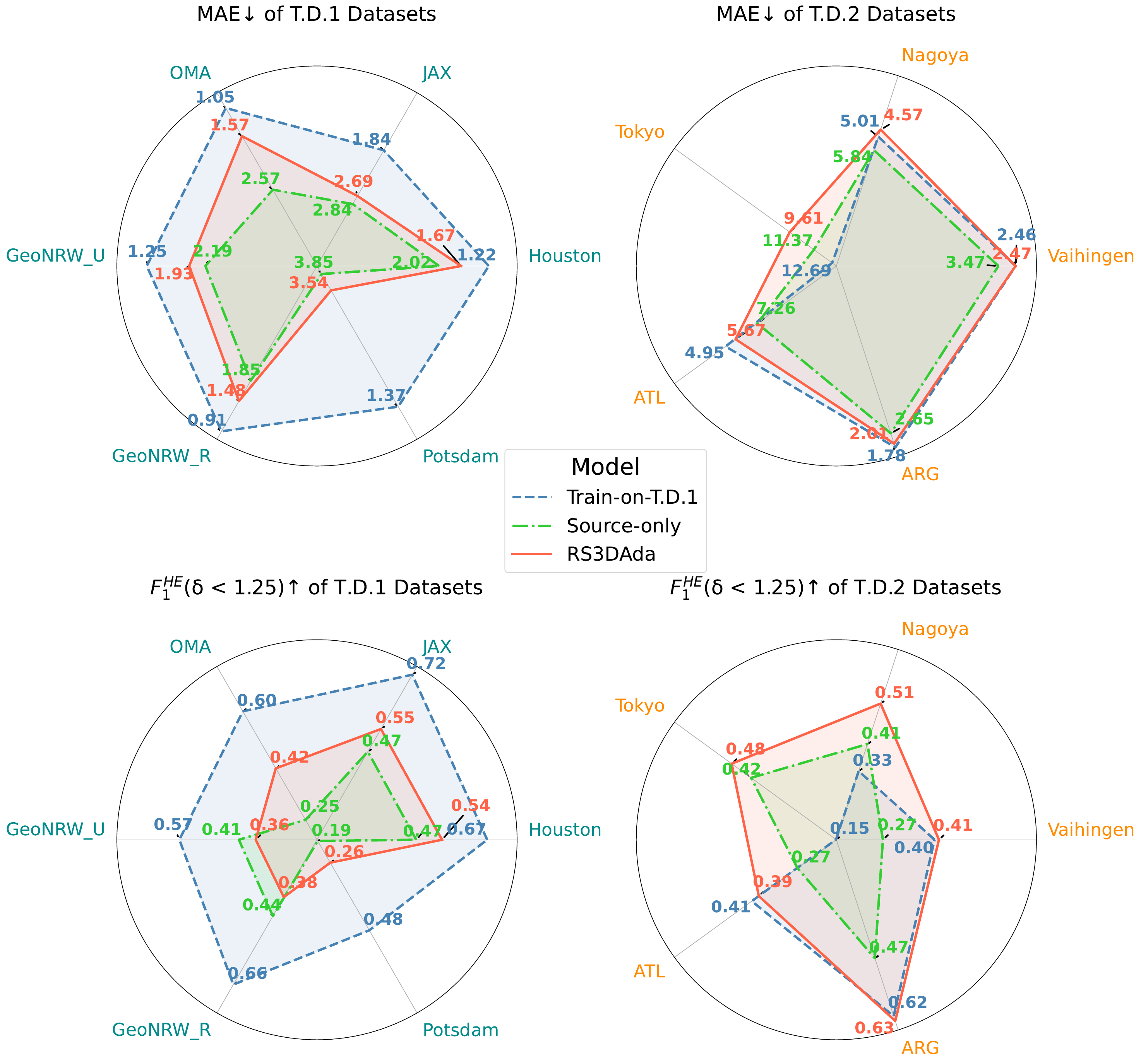}
    \caption{Results of RS3DAda height estimation branch for each dataset.}
    \label{fig:mtda_composite}
  \end{minipage}
\end{figure}

\noindent \textbf{Heigh Estimation Branch.} Table~\ref{tab:mtda_overall} shows the height estimation results for our RS3DAda model. "Whole" refers to the entire image, and "High" focuses on targets above 3 meters, such as trees and buildings. Using DINOv2 \cite{oquab2023dinov2} and DPT \cite{ranftl2021vision} models, RS3DAda reduces the average MAE by 0.409 meters over the source-only approach across six datasets in \textit{Target Domain 1}, though it is still exceeded by the models trained directly on these datasets. In \textit{Target Domain 2}, RS3DAda outperforms models trained on real-world data, indicating its strong generalization under challenging scenarios, featuring diverse geographic regions and complex terrain characteristics. Figure~\ref{fig:mtda_composite} aligns with these results, showing RS3DAda's improvements on each dataset in \textit{Target Domain 1} and \textit{Target Domain 2}. This demonstrates SynRS3D's potential and the effectiveness of RS3DAda. However, the gap in \textit{Target Domain 1} highlights the ongoing need to bridge the synthetic-to-real data gap, providing a benchmark for future UDA algorithm development in height estimation tasks.

\noindent \textbf{Land Cover Mapping Branch.}
\begin{table}[h]
\caption{Results of the RS3DAda land cover mapping branch on the OEM \cite{xia2023openearthmap} dataset. All models are implemented with DINOv2 \cite{oquab2023dinov2} and DPT \cite{ranftl2021vision}.}
\label{tab:mtda_oem}
\centering
\begin{adjustbox}{width=\textwidth,center}
\begin{tabular}{l|cccccccc|c}
\toprule
Model & Bareland & Rangeland & Developed & Road & Tree & Water & Agriculture & Buildings & \textbf{mIoU} \\
\midrule
Source-only & 8.69 & 37.95 & \textbf{22.54} & \textbf{49.05} & 60.16 & 46.64 & 35.40 & \textbf{65.19} & 40.70 \\
DAFormer \cite{hoyer2022daformer} & 12.54 & 41.16 & 10.88 & 43.88 & \textbf{62.56} & \textbf{77.55} & 62.62 & 59.10 & 46.29 \\
\textbf{RS3DAda} & \textbf{19.92} & \textbf{47.61} & 18.41 & 44.06 & 61.04 & 71.66 & \textbf{63.73} & 59.42 & \textbf{48.23} \\
\rowcolor{gray!20} \textcolor{gray!80}{Train-on-OEM} & \textcolor{gray!80}{50.04} & \textcolor{gray!80}{59.10} & \textcolor{gray!80}{58.18} & \textcolor{gray!80}{65.39} & \textcolor{gray!80}{73.07} & \textcolor{gray!80}{83.65} & \textcolor{gray!80}{76.36} & \textcolor{gray!80}{80.88} & \textcolor{gray!80}{68.34} \\
\bottomrule
\end{tabular}
\end{adjustbox}
\end{table}
Table~\ref{tab:mtda_oem} presents the results of the RS3DAda model for the land cover mapping branch, evaluated using the OEM dataset. As shown, the RS3DAda method surpasses DAFormer by \(1.94\) in mIoU, indicating that the height branch positively impacts the land cover mapping performance. However, there remains a gap of \(20.11\) in mIoU compared to the Oracle model, suggesting significant room for improvement. Future research can build upon our method to make further advancements.

\begin{figure}[htbp]
  \begin{minipage}[b]{0.5\linewidth}
    \centering
    \includegraphics[width=\linewidth]{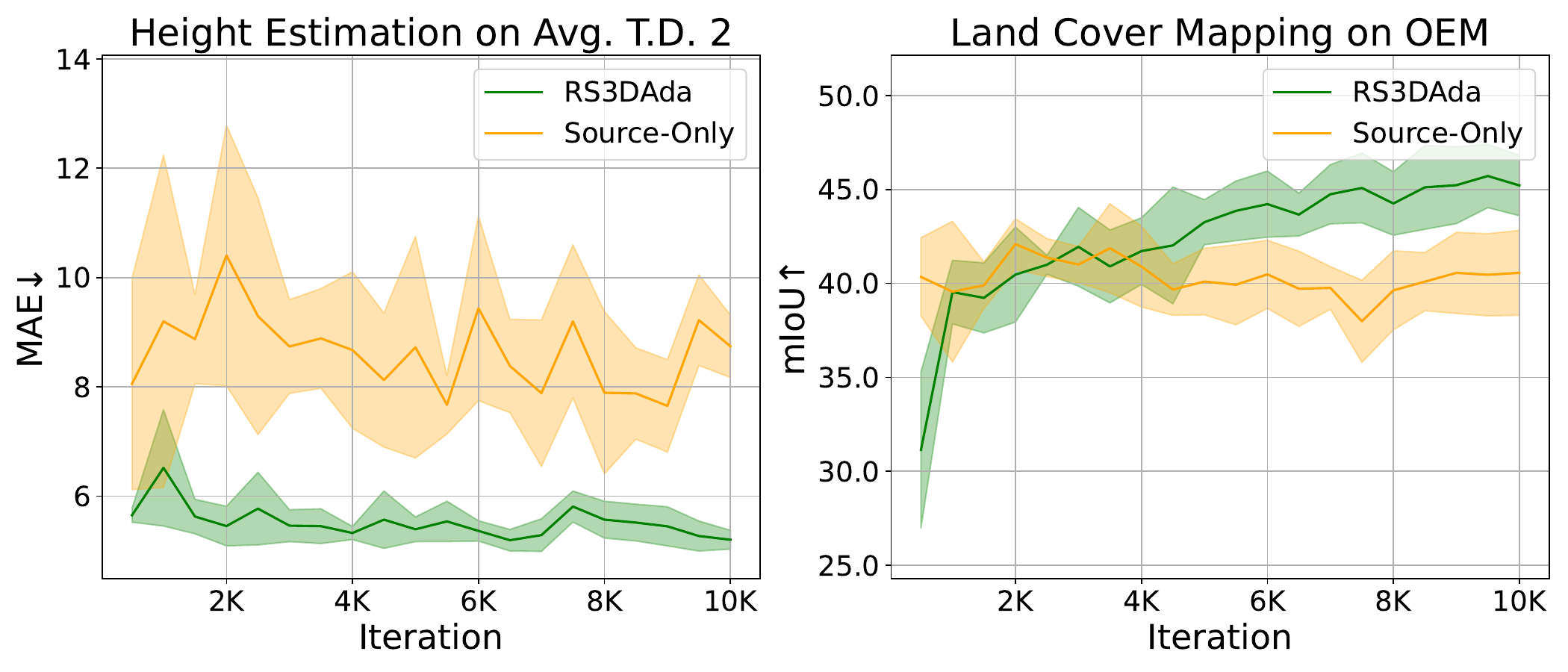}
    \caption{Performance of SynRS3D at the beginning of training for height estimation and land cover mapping branches, with and without the use of RS3DAda.}
    \label{fig:mtda_stability}
  \end{minipage}
  \hspace{0.03\linewidth}
  \begin{minipage}[b]{0.45\linewidth}
    \noindent \textbf{Stabilizing Training on SynRS3D.} RS3DAda can regularize the training of synthetic data to prevent rapid overfitting at the beginning of the training, corresponding to the green section in Figure~\ref{fig:mtda_stability}. Without RS3DAda, the model's evaluation results on the target domain fluctuate wildly during training in both height estimation and land cover mapping branches. This instability can lead to unreliable performance and poor generalization. RS3DAda ensures more consistent training, resulting in better model accuracy and stability.
  \end{minipage}
\end{figure}

\begin{figure}[h!]
  \centering
  \begin{minipage}[h!]{0.51\linewidth} 
    \centering
    \captionof{table}{Comparison of RS3DAda with existing UDA methods AdaSeg and DADA. Supervision types: H for height maps, L for land cover labels. T.D.1 represents Target Domain 1, and T.D.2 represents Target Domain 2. V: Vaihingen \cite{rottensteiner2012isprs}; P: Potsdam \cite{rottensteiner2012isprs}; J: JAX \cite{dfc2019}; O: OMA \cite{dfc2019}. All models are implemented with DeepLabv2 and ResNet101. The datasets used for Tranin-on-Real are T.D.1 and OEM respectively.}
    \label{tab:deeplab_mtda}
    \begin{adjustbox}{width=\textwidth,center}
      \begin{tabular}{l|c|cc|cc}
      \toprule
      Model & Supervision & \multicolumn{2}{c}{Height Estimation (MAE) $\downarrow$} & \multicolumn{2}{c}{Land Cover Mapping (mIoU) $\uparrow$} \\
      \cmidrule(lr){3-4} \cmidrule(lr){5-6}
      & & {Avg. T.D.1} & {Avg. T.D.2} & {OEM \cite{xia2023openearthmap}} & {Avg. (V+P+J+O)} \\
      \midrule
      {Source-Only} & H, L & 3.911 & 7.419 & 17.42 & 39.61 \\
      {AdaptSeg \cite{tsai2018learning}} & L & - & - & 20.06 & 40.00 \\
      {DADA \cite{Vu_2019_ICCV}} & H+L & 3.615 & 6.997 & 21.24 & 46.44 \\
      \textbf{RS3DAda} & H+L & \textbf{3.275} & \textbf{6.708} & \textbf{22.55} & \textbf{47.28} \\
      \rowcolor{gray!20}\textcolor{gray!80}{{Train-on-Real}} & \textcolor{gray!80}{H, L} & \textcolor{gray!80}{1.859} & \textcolor{gray!80}{6.639} & \textcolor{gray!80}{64.54} & \textcolor{gray!80}{53.12} \\
      \bottomrule
      \end{tabular}
    \end{adjustbox}
  \end{minipage}
  \hspace{0.03\linewidth}
  \begin{minipage}[h!]{0.44\linewidth} 
    \noindent \textbf{Comparison with Existing UDA.} We re-implemented AdaptSeg \cite{tsai2018learning}, DADA \cite{Vu_2019_ICCV} and RS3DAda using DeepLabv2 \cite{chen2017deeplab} and ResNet101 \cite{he2016deep} for fair comparison. Table~\ref{tab:deeplab_mtda} shows that RS3DAda outperformed both AdaptSeg and DADA in height estimation and land cover mapping tasks. Notably, when using weaker network architectures, the CNN encoder pre-trained on ImageNet fails to provide reliable RS image features and high accuracy for the ground category. As a result, the Feature Constraint and Ground-Guided Pseudo-Label Refinement in our RS3DAda cannot achieve their maximum effectiveness.
  \end{minipage}
\end{figure}

\section{Discussion}
\label{sec:discuss}
We acknowledge limitations in our SynRS3D dataset and RS3DAda methodology. A key challenge is the appearance gap between SynRS3D and real data, especially without UDA technology. RS3DAda is specific to RS and synthetic-to-real multi-task UDA, limiting its application to non-RS scenarios and other domain adaptation types. By making our dataset, methods, and model weights publicly available, we aim to advance RS applications that enhance human well-being and environmental sustainability. We emphasize the importance of using our data and model responsibly and caution against applications that could infringe on privacy or contravene human rights.

\section{Conclusion}
In this study, we introduced SynRS3D, the world's largest synthetic RS dataset, and RS3DAda, a UDA method. These tools aim to bridge the research gap by enabling global 3D semantic reconstruction from single-view RS images. Our evaluations on multiple public datasets show that SynRS3D and RS3DAda can boost the utility of synthetic data for RS research. Although a performance gap remains compared to real data, our work sets a benchmark for 3D semantic reconstruction using synthetic data from monocular RS images, highlighting the potential of SynRS3D and RS3DAda for future research and applications.

\bibliographystyle{plain}
\small\bibliography{main}

\begin{thebibliography}{100}

\bibitem{abramov2020keep}
Alexey Abramov, Christopher Bayer, and Claudio Heller.
\newblock Keep it simple: Image statistics matching for domain adaptation.
\newblock {\em arXiv preprint arXiv:2005.12551}, 2020.

\bibitem{ameri2002high}
B~Ameri, N~Goldstein, H~Wehn, A~Moshkovitz, and H~Zwick.
\newblock High resolution digital surface model (dsm) generation using multi-view multi-frame digital airborne images.
\newblock {\em INTERNATIONAL ARCHIVES OF PHOTOGRAMMETRY REMOTE SENSING AND SPATIAL INFORMATION SCIENCES}, 34(4):419--424, 2002.

\bibitem{atapour2018real}
Amir Atapour-Abarghouei and Toby~P Breckon.
\newblock Real-time monocular depth estimation using synthetic data with domain adaptation via image style transfer.
\newblock In {\em Proceedings of the IEEE conference on computer vision and pattern recognition}, pages 2800--2810, 2018.

\bibitem{baier2021synthesizing}
Gerald Baier, Antonin Deschemps, Michael Schmitt, and Naoto Yokoya.
\newblock Synthesizing optical and sar imagery from land cover maps and auxiliary raster data.
\newblock {\em IEEE Transactions on Geoscience and Remote Sensing}, 60:1--12, 2021.

\bibitem{bandara2022transformer}
Wele Gedara~Chaminda Bandara and Vishal~M Patel.
\newblock A transformer-based siamese network for change detection.
\newblock In {\em IGARSS 2022-2022 IEEE International Geoscience and Remote Sensing Symposium}, pages 207--210. IEEE, 2022.

\bibitem{bourdis2011constrained}
Nicolas Bourdis, Denis Marraud, and Hichem Sahbi.
\newblock Constrained optical flow for aerial image change detection.
\newblock In {\em 2011 IEEE international geoscience and remote sensing symposium}, pages 4176--4179. IEEE, 2011.

\bibitem{butler2012naturalistic}
Daniel~J Butler, Jonas Wulff, Garrett~B Stanley, and Michael~J Black.
\newblock A naturalistic open source movie for optical flow evaluation.
\newblock In {\em Computer Vision--ECCV 2012: 12th European Conference on Computer Vision, Florence, Italy, October 7-13, 2012, Proceedings, Part VI 12}, pages 611--625. Springer, 2012.

\bibitem{chen2020spatial}
Hao Chen and Zhenwei Shi.
\newblock A spatial-temporal attention-based method and a new dataset for remote sensing image change detection.
\newblock {\em Remote Sensing}, 12(10):1662, 2020.

\bibitem{chen2024changemamba}
Hongruixuan Chen, Jian Song, Chengxi Han, Junshi Xia, and Naoto Yokoya.
\newblock Changemamba: Remote sensing change detection with spatio-temporal state space model.
\newblock 2024.

\bibitem{chen2017deeplab}
Liang-Chieh Chen, George Papandreou, Iasonas Kokkinos, Kevin Murphy, and Alan~L Yuille.
\newblock Deeplab: Semantic image segmentation with deep convolutional nets, atrous convolution, and fully connected crfs.
\newblock {\em IEEE transactions on pattern analysis and machine intelligence}, 40(4):834--848, 2017.

\bibitem{chen2019crdoco}
Yun-Chun Chen, Yen-Yu Lin, Ming-Hsuan Yang, and Jia-Bin Huang.
\newblock Crdoco: Pixel-level domain transfer with cross-domain consistency.
\newblock In {\em Proceedings of the IEEE/CVF conference on computer vision and pattern recognition}, pages 1791--1800, 2019.

\bibitem{ogc}
Gordon Christie, Kevin Foster, Shea Hagstrom, Gregory~D Hager, and Myron~Z Brown.
\newblock Single view geocentric pose in the wild.
\newblock In {\em Proceedings of the IEEE/CVF Conference on Computer Vision and Pattern Recognition}, pages 1162--1171, 2021.

\bibitem{blender}
Blender~Online Community.
\newblock {\em Blender - a 3D modelling and rendering package}.
\newblock Blender Foundation, Stichting Blender Foundation, Amsterdam, 2018.

\bibitem{aw3d2018}
NTT~DATA Corporation and Inc. DigitalGlobe.
\newblock Aw3d high-resolution dataset.
\newblock End User License Agreement, 2018.
\newblock Available from NTT DATA Corporation and DigitalGlobe, Inc.

\bibitem{demir2018deepglobe}
Ilke Demir, Krzysztof Koperski, David Lindenbaum, Guan Pang, Jing Huang, Saikat Basu, Forest Hughes, Devis Tuia, and Ramesh Raskar.
\newblock Deepglobe 2018: A challenge to parse the earth through satellite images.
\newblock In {\em Proceedings of the IEEE Conference on Computer Vision and Pattern Recognition Workshops}, pages 172--181, 2018.

\bibitem{dosovitskiy2020image}
Alexey Dosovitskiy, Lucas Beyer, Alexander Kolesnikov, Dirk Weissenborn, Xiaohua Zhai, Thomas Unterthiner, Mostafa Dehghani, Matthias Minderer, Georg Heigold, Sylvain Gelly, et~al.
\newblock An image is worth 16x16 words: Transformers for image recognition at scale.
\newblock {\em arXiv preprint arXiv:2010.11929}, 2020.

\bibitem{eigen2014depth}
David Eigen, Christian Puhrsch, and Rob Fergus.
\newblock Depth map prediction from a single image using a multi-scale deep network.
\newblock {\em Advances in neural information processing systems}, 27, 2014.

\bibitem{multiview3d_2}
Massimiliano Favalli, Alessandro Fornaciai, Ilaria Isola, Simone Tarquini, and Luca Nannipieri.
\newblock Multiview 3d reconstruction in geosciences.
\newblock {\em Computers \& Geosciences}, 44:168--176, 2012.

\bibitem{multiview3d_4}
Jian Gao, Jin Liu, and Shunping Ji.
\newblock A general deep learning based framework for 3d reconstruction from multi-view stereo satellite images.
\newblock {\em ISPRS Journal of Photogrammetry and Remote Sensing}, 195:446--461, 2023.

\bibitem{joint_height}
Zhi Gao, Wenbo Sun, Yao Lu, Yichen Zhang, Weiwei Song, Yongjun Zhang, and Ruifang Zhai.
\newblock Joint learning of semantic segmentation and height estimation for remote sensing image leveraging contrastive learning.
\newblock {\em IEEE Transactions on Geoscience and Remote Sensing}, 2023.

\bibitem{img2dsm}
Pedram Ghamisi and Naoto Yokoya.
\newblock Img2dsm: Height simulation from single imagery using conditional generative adversarial net.
\newblock {\em IEEE Geoscience and Remote Sensing Letters}, 15(5):794--798, 2018.

\bibitem{gong2021dlow}
Rui Gong, Wen Li, Yuhua Chen, Dengxin Dai, and Luc Van~Gool.
\newblock Dlow: Domain flow and applications.
\newblock {\em International Journal of Computer Vision}, 129(10):2865--2888, 2021.

\bibitem{goodfellow2014generative}
Ian Goodfellow, Jean Pouget-Abadie, Mehdi Mirza, Bing Xu, David Warde-Farley, Sherjil Ozair, Aaron Courville, and Yoshua Bengio.
\newblock Generative adversarial nets.
\newblock In {\em Advances in Neural Information Processing Systems}, pages 2672--2680, 2014.

\bibitem{gordon2020learning}
C~Gordon, R~Rene Rai~Munoz Abujder, K~Foster, S~Hagstrom, GD~Hager, and MZ~Brown.
\newblock Learning geocentric object pose in oblique monocular images.
\newblock In {\em Proceedings of the IEEE Computer Society Conference on Computer Vision and Pattern Recognition}, 2020.

\bibitem{han2020state}
Yilong Han, Shugen Wang, Danchao Gong, Yue Wang, and X~Ma.
\newblock State of the art in digital surface modelling from multi-view high-resolution satellite images.
\newblock {\em ISPRS Annals of the Photogrammetry, Remote Sensing and Spatial Information Sciences}, 2:351--356, 2020.

\bibitem{handa2016scenenet}
Ankur Handa, Viorica P{\u{a}}tr{\u{a}}ucean, Simon Stent, and Roberto Cipolla.
\newblock Scenenet: An annotated model generator for indoor scene understanding.
\newblock In {\em 2016 IEEE International Conference on Robotics and Automation (ICRA)}, pages 5737--5743. IEEE, 2016.

\bibitem{he2016deep}
Kaiming He, Xiangyu Zhang, Shaoqing Ren, and Jian Sun.
\newblock Deep residual learning for image recognition.
\newblock In {\em Proceedings of the IEEE conference on computer vision and pattern recognition}, pages 770--778, 2016.

\bibitem{hermosilla2011evaluation}
Txomin Hermosilla, Luis~A Ruiz, Jorge~A Recio, and Javier Estornell.
\newblock Evaluation of automatic building detection approaches combining high resolution images and lidar data.
\newblock {\em Remote Sensing}, 3(6):1188--1210, 2011.

\bibitem{ho2020denoising}
Jonathan Ho, Ajay Jain, and Pieter Abbeel.
\newblock Denoising diffusion probabilistic models.
\newblock {\em Advances in Neural Information Processing Systems}, 33:6840--6851, 2020.

\bibitem{hoffman2018cycada}
Judy Hoffman, Eric Tzeng, Taesung Park, Jun-Yan Zhu, Phillip Isola, Kate Saenko, Alexei Efros, and Trevor Darrell.
\newblock Cycada: Cycle-consistent adversarial domain adaptation.
\newblock In {\em International conference on machine learning}, pages 1989--1998. Pmlr, 2018.

\bibitem{hoffman2016fcns}
Judy Hoffman, Dequan Wang, Fisher Yu, and Trevor Darrell.
\newblock Fcns in the wild: Pixel-level adversarial and constraint-based adaptation.
\newblock {\em arXiv preprint arXiv:1612.02649}, 2016.

\bibitem{hong2023cross}
Danfeng Hong, Bing Zhang, Hao Li, Yuxuan Li, Jing Yao, Chenyu Li, Martin Werner, Jocelyn Chanussot, Alexander Zipf, and Xiao~Xiang Zhu.
\newblock Cross-city matters: A multimodal remote sensing benchmark dataset for cross-city semantic segmentation using high-resolution domain adaptation networks.
\newblock {\em Remote Sensing of Environment}, 299:113856, 2023.

\bibitem{hoyer2022daformer}
Lukas Hoyer, Dengxin Dai, and Luc Van~Gool.
\newblock Daformer: Improving network architectures and training strategies for domain-adaptive semantic segmentation.
\newblock In {\em Proceedings of the IEEE/CVF Conference on Computer Vision and Pattern Recognition}, pages 9924--9935, 2022.

\bibitem{hoyer2022hrda}
Lukas Hoyer, Dengxin Dai, and Luc Van~Gool.
\newblock Hrda: Context-aware high-resolution domain-adaptive semantic segmentation.
\newblock In {\em European conference on computer vision}, pages 372--391. Springer, 2022.

\bibitem{hoyer2023mic}
Lukas Hoyer, Dengxin Dai, Haoran Wang, and Luc Van~Gool.
\newblock Mic: Masked image consistency for context-enhanced domain adaptation.
\newblock In {\em Proceedings of the IEEE/CVF conference on computer vision and pattern recognition}, pages 11721--11732, 2023.

\bibitem{multiview3d_3}
Zhihua Hu, Yaolin Hou, Pengjie Tao, and Jie Shan.
\newblock Imgtr: Image-triangle based multi-view 3d reconstruction for urban scenes.
\newblock {\em ISPRS Journal of Photogrammetry and Remote Sensing}, 181:191--204, 2021.

\bibitem{iqbal2020weakly}
Javed Iqbal and Mohsen Ali.
\newblock Weakly-supervised domain adaptation for built-up region segmentation in aerial and satellite imagery.
\newblock {\em ISPRS Journal of Photogrammetry and Remote Sensing}, 167:263--275, 2020.

\bibitem{ji2018fully}
Shunping Ji, Shiqing Wei, and Meng Lu.
\newblock Fully convolutional networks for multisource building extraction from an open aerial and satellite imagery data set.
\newblock {\em IEEE Transactions on Geoscience and Remote Sensing}, 57(1):574--586, 2018.

\bibitem{kim2018procedural}
Joon-Seok Kim, Hamdi Kavak, and Andrew Crooks.
\newblock Procedural city generation beyond game development.
\newblock {\em SIGSPATIAL Special}, 10(2):34--41, 2018.

\bibitem{kong2020synthinel}
Fanjie Kong, Bohao Huang, Kyle Bradbury, and Jordan Malof.
\newblock The synthinel-1 dataset: A collection of high resolution synthetic overhead imagery for building segmentation.
\newblock In {\em Proceedings of the IEEE/CVF winter conference on applications of computer vision}, pages 1814--1823, 2020.

\bibitem{joint_height3}
Saket Kunwar.
\newblock U-net ensemble for semantic and height estimation using coarse-map initialization.
\newblock In {\em IGARSS 2019-2019 IEEE International Geoscience and Remote Sensing Symposium}, pages 4959--4962. IEEE, 2019.

\bibitem{dfc2019}
Bertrand Le~Saux, Naoto Yokoya, Ronny H{\"a}nsch, and Myron Brown.
\newblock 2019 ieee grss data fusion contest: large-scale semantic 3d reconstruction.
\newblock {\em IEEE Geoscience and Remote Sensing Magazine (GRSM)}, 7(4):33--36, 2019.

\bibitem{lee2013pseudo}
Dong-Hyun Lee et~al.
\newblock Pseudo-label: The simple and efficient semi-supervised learning method for deep neural networks.
\newblock In {\em Workshop on challenges in representation learning, ICML}, volume~3, page 896. Atlanta, 2013.

\bibitem{multiview3d_6}
Matthew~J Leotta, Chengjiang Long, Bastien Jacquet, Matthieu Zins, Dan Lipsa, Jie Shan, Bo~Xu, Zhixin Li, Xu~Zhang, Shih-Fu Chang, et~al.
\newblock Urban semantic 3d reconstruction from multiview satellite imagery.
\newblock In {\em Proceedings of the IEEE/CVF Conference on Computer Vision and Pattern Recognition Workshops}, pages 0--0, 2019.

\bibitem{building_height}
Qingyu Li, Lichao Mou, Yuansheng Hua, Yilei Shi, Sining Chen, Yao Sun, and Xiao~Xiang Zhu.
\newblock 3dcentripetalnet: Building height retrieval from monocular remote sensing imagery.
\newblock {\em International Journal of Applied Earth Observation and Geoinformation}, 120:103311, 2023.

\bibitem{li20193d}
Suo Li, Zhanyu Zhu, Haipeng Wang, and Feng Xu.
\newblock 3d virtual urban scene reconstruction from a single optical remote sensing image.
\newblock {\em IEEE Access}, 7:68305--68315, 2019.

\bibitem{li2020high}
Wang Li, Zheng Niu, Rong Shang, Yuchu Qin, Li~Wang, and Hanyue Chen.
\newblock High-resolution mapping of forest canopy height using machine learning by coupling icesat-2 lidar with sentinel-1, sentinel-2 and landsat-8 data.
\newblock {\em International Journal of Applied Earth Observation and Geoinformation}, 92:102163, 2020.

\bibitem{3dbuilidng1}
Weijia Li, Lingxuan Meng, Jinwang Wang, Conghui He, Gui-Song Xia, and Dahua Lin.
\newblock 3d building reconstruction from monocular remote sensing images.
\newblock In {\em Proceedings of the IEEE/CVF International Conference on Computer Vision}, pages 12548--12557, 2021.

\bibitem{3dbuilding2}
Weijia Li, Haote Yang, Zhenghao Hu, Juepeng Zheng, Gui-Song Xia, and Conghui He.
\newblock 3d building reconstruction from monocular remote sensing images with multi-level supervisions.
\newblock {\em arXiv preprint arXiv:2404.04823}, 2024.

\bibitem{building_height2}
Xiang Li, Mingyang Wang, and Yi~Fang.
\newblock Height estimation from single aerial images using a deep ordinal regression network.
\newblock {\em IEEE Geoscience and Remote Sensing Letters}, 19:1--5, 2020.

\bibitem{li2023synergistical}
Xin Li, Feng Xu, Fan Liu, Xin Lyu, Yao Tong, Zhennan Xu, and Jun Zhou.
\newblock A synergistical attention model for semantic segmentation of remote sensing images.
\newblock {\em IEEE Transactions on Geoscience and Remote Sensing}, 61:1--16, 2023.

\bibitem{lian2019constructing}
Qing Lian, Fengmao Lv, Lixin Duan, and Boqing Gong.
\newblock Constructing self-motivated pyramid curriculums for cross-domain semantic segmentation: A non-adversarial approach.
\newblock In {\em Proceedings of the IEEE/CVF International Conference on Computer Vision}, pages 6758--6767, 2019.

\bibitem{liu2023deep}
Jin Liu, Jian Gao, Shunping Ji, Chang Zeng, Shaoyi Zhang, and JianYa Gong.
\newblock Deep learning based multi-view stereo matching and 3d scene reconstruction from oblique aerial images.
\newblock {\em ISPRS Journal of Photogrammetry and Remote Sensing}, 204:42--60, 2023.

\bibitem{liu2024source}
Weixing Liu, Jun Liu, Xin Su, Han Nie, and Bin Luo.
\newblock Source-free domain adaptive object detection in remote sensing images.
\newblock {\em arXiv preprint arXiv:2401.17916}, 2024.

\bibitem{liu2020building}
Yi~Liu, Chao Pang, Zongqian Zhan, Xiaomeng Zhang, and Xue Yang.
\newblock Building change detection for remote sensing images using a dual-task constrained deep siamese convolutional network model.
\newblock {\em IEEE Geoscience and Remote Sensing Letters}, 18(5):811--815, 2020.

\bibitem{liu2023rethinking}
Yuheng Liu, Yifan Zhang, Ye~Wang, and Shaohui Mei.
\newblock Rethinking transformers for semantic segmentation of remote sensing images.
\newblock {\em IEEE Transactions on Geoscience and Remote Sensing}, 2023.

\bibitem{lopez2023desc}
Adrian Lopez-Rodriguez and Krystian Mikolajczyk.
\newblock Desc: Domain adaptation for depth estimation via semantic consistency.
\newblock {\em International Journal of Computer Vision}, 131(3):752--771, 2023.

\bibitem{lu2019multisource}
Xiaoqiang Lu, Tengfei Gong, and Xiangtao Zheng.
\newblock Multisource compensation network for remote sensing cross-domain scene classification.
\newblock {\em IEEE Transactions on Geoscience and Remote Sensing}, 58(4):2504--2515, 2019.

\bibitem{luo2019taking}
Yawei Luo, Liang Zheng, Tao Guan, Junqing Yu, and Yi~Yang.
\newblock Taking a closer look at domain shift: Category-level adversaries for semantics consistent domain adaptation.
\newblock In {\em Proceedings of the IEEE/CVF conference on computer vision and pattern recognition}, pages 2507--2516, 2019.

\bibitem{mahphood2019dense}
A~Mahphood, H~Arefi, A~Hosseininaveh, and AA~Naeini.
\newblock Dense multi-view image matching for dsm generation from satellite images.
\newblock {\em The International Archives of the Photogrammetry, Remote Sensing and Spatial Information Sciences}, 42:709--715, 2019.

\bibitem{3dbuilding3}
Yongqiang Mao, Kaiqiang Chen, Liangjin Zhao, Wei Chen, Deke Tang, Wenjie Liu, Zhirui Wang, Wenhui Diao, Xian Sun, and Kun Fu.
\newblock Elevation estimation-driven building 3d reconstruction from single-view remote sensing imagery.
\newblock {\em IEEE Transactions on Geoscience and Remote Sensing}, 2023.

\bibitem{mayer2016large}
Nikolaus Mayer, Eddy Ilg, Philip Hausser, Philipp Fischer, Daniel Cremers, Alexey Dosovitskiy, and Thomas Brox.
\newblock A large dataset to train convolutional networks for disparity, optical flow, and scene flow estimation.
\newblock In {\em Proceedings of the IEEE conference on computer vision and pattern recognition}, pages 4040--4048, 2016.

\bibitem{mei2020instance}
Ke~Mei, Chuang Zhu, Jiaqi Zou, and Shanghang Zhang.
\newblock Instance adaptive self-training for unsupervised domain adaptation.
\newblock In {\em Computer Vision--ECCV 2020: 16th European Conference, Glasgow, UK, August 23--28, 2020, Proceedings, Part XXVI 16}, pages 415--430. Springer, 2020.

\bibitem{muller2006procedural}
Pascal M{\"u}ller, Peter Wonka, Simon Haegler, Andreas Ulmer, and Luc Van~Gool.
\newblock Procedural modeling of buildings.
\newblock In {\em ACM SIGGRAPH 2006 Papers}, pages 614--623. 2006.

\bibitem{musgrave1989synthesis}
F~Kenton Musgrave, Craig~E Kolb, and Robert~S Mace.
\newblock The synthesis and rendering of eroded fractal terrains.
\newblock {\em ACM Siggraph Computer Graphics}, 23(3):41--50, 1989.

\bibitem{olsson2021classmix}
Viktor Olsson, Wilhelm Tranheden, Juliano Pinto, and Lennart Svensson.
\newblock Classmix: Segmentation-based data augmentation for semi-supervised learning.
\newblock In {\em Proceedings of the IEEE/CVF winter conference on applications of computer vision}, pages 1369--1378, 2021.

\bibitem{openai2023gpt4}
OpenAI.
\newblock Gpt-4 technical report, 2023.

\bibitem{oquab2023dinov2}
Maxime Oquab, Timoth{\'e}e Darcet, Th{\'e}o Moutakanni, Huy Vo, Marc Szafraniec, Vasil Khalidov, Pierre Fernandez, Daniel Haziza, Francisco Massa, Alaaeldin El-Nouby, et~al.
\newblock Dinov2: Learning robust visual features without supervision.
\newblock {\em arXiv preprint arXiv:2304.07193}, 2023.

\bibitem{othman2017domain}
Esam Othman, Yakoub Bazi, Farid Melgani, Haikel Alhichri, Naif Alajlan, and Mansour Zuair.
\newblock Domain adaptation network for cross-scene classification.
\newblock {\em IEEE Transactions on Geoscience and Remote Sensing}, 55(8):4441--4456, 2017.

\bibitem{dfc2023}
Claudio Persello, Ronny Hänsch, Gemine Vivone, Kaiqiang Chen, Zhiyuan Yan, Deke Tang, Hai Huang, Michael Schmitt, and Xian Sun.
\newblock 2023 ieee grss data fusion contest: Large-scale fine-grained building classification for semantic urban reconstruction [technical committees].
\newblock {\em IEEE Geoscience and Remote Sensing Magazine}, 11(1):94--97, 2023.

\bibitem{ranftl2021vision}
Ren{\'e} Ranftl, Alexey Bochkovskiy, and Vladlen Koltun.
\newblock Vision transformers for dense prediction.
\newblock In {\em Proceedings of the IEEE/CVF international conference on computer vision}, pages 12179--12188, 2021.

\bibitem{reyes2022syntcities}
Mario~Fuentes Reyes, Pablo d'Angelo, and Friedrich Fraundorfer.
\newblock Syntcities: A large synthetic remote sensing dataset for disparity estimation.
\newblock {\em IEEE Journal of Selected Topics in Applied Earth Observations and Remote Sensing}, 15:10087--10098, 2022.

\bibitem{reyes20232d}
Mario~Fuentes Reyes, Yuxing Xie, Xiangtian Yuan, Pablo d’Angelo, Franz Kurz, Daniele Cerra, and Jiaojiao Tian.
\newblock A 2d/3d multimodal data simulation approach with applications on urban semantic segmentation, building extraction and change detection.
\newblock {\em ISPRS Journal of Photogrammetry and Remote Sensing}, 205:74--97, 2023.

\bibitem{richter2016playing}
Stephan~R Richter, Vibhav Vineet, Stefan Roth, and Vladlen Koltun.
\newblock Playing for data: Ground truth from computer games.
\newblock In {\em Computer Vision--ECCV 2016: 14th European Conference, Amsterdam, The Netherlands, October 11-14, 2016, Proceedings, Part II 14}, pages 102--118. Springer, 2016.

\bibitem{rombach2022high}
Robin Rombach, Andreas Blattmann, Dominik Lorenz, Patrick Esser, and Bj{\"o}rn Ommer.
\newblock High-resolution image synthesis with latent diffusion models.
\newblock In {\em Proceedings of the IEEE/CVF conference on computer vision and pattern recognition}, pages 10684--10695, 2022.

\bibitem{ros2016synthia}
German Ros, Laura Sellart, Joanna Materzynska, David Vazquez, and Antonio~M Lopez.
\newblock The synthia dataset: A large collection of synthetic images for semantic segmentation of urban scenes.
\newblock In {\em Proceedings of the IEEE conference on computer vision and pattern recognition}, pages 3234--3243, 2016.

\bibitem{rottensteiner2012isprs}
Franz Rottensteiner, Gunho Sohn, Jaewook Jung, Markus Gerke, Caroline Baillard, Sebastien Benitez, and Uwe Breitkopf.
\newblock The isprs benchmark on urban object classification and 3d building reconstruction.
\newblock {\em ISPRS Annals of the Photogrammetry, Remote Sensing and Spatial Information Sciences; I-3}, 1(1):293--298, 2012.

\bibitem{multiview3d_5}
Ewelina Rupnik, Marc Pierrot-Deseilligny, and Arthur Delorme.
\newblock 3d reconstruction from multi-view vhr-satellite images in micmac.
\newblock {\em ISPRS Journal of Photogrammetry and Remote Sensing}, 139:201--211, 2018.

\bibitem{sakaridis2018model}
Christos Sakaridis, Dengxin Dai, Simon Hecker, and Luc Van~Gool.
\newblock Model adaptation with synthetic and real data for semantic dense foggy scene understanding.
\newblock In {\em Proceedings of the european conference on computer vision (ECCV)}, pages 687--704, 2018.

\bibitem{schmitt2023there}
Michael Schmitt, Seyed~Ali Ahmadi, Yonghao Xu, G{\"u}l{\c{s}}en Ta{\c{s}}k{\i}n, Ujjwal Verma, Francescopaolo Sica, and Ronny H{\"a}nsch.
\newblock There are no data like more data: Datasets for deep learning in earth observation.
\newblock {\em IEEE Geoscience and Remote Sensing Magazine}, 2023.

\bibitem{shermeyer2021rareplanes}
Jacob Shermeyer, Thomas Hossler, Adam Van~Etten, Daniel Hogan, Ryan Lewis, and Daeil Kim.
\newblock Rareplanes: Synthetic data takes flight.
\newblock In {\em Proceedings of the IEEE/CVF Winter Conference on Applications of Computer Vision}, pages 207--217, 2021.

\bibitem{sohn2004extraction}
Gunho Sohn and Ian~J Dowman.
\newblock Extraction of buildings from high resolution satellite data and lidar.
\newblock In {\em XX ISPRS CONGRESS}, 2004.

\bibitem{song2024syntheworld}
Jian Song, Hongruixuan Chen, and Naoto Yokoya.
\newblock Syntheworld: A large-scale synthetic dataset for land cover mapping and building change detection.
\newblock In {\em Proceedings of the IEEE/CVF Winter Conference on Applications of Computer Vision}, pages 8287--8296, 2024.

\bibitem{joint_height4}
Shivangi Srivastava, Michele Volpi, and Devis Tuia.
\newblock Joint height estimation and semantic labeling of monocular aerial images with cnns.
\newblock In {\em 2017 IEEE International Geoscience and Remote Sensing Symposium (IGARSS)}, pages 5173--5176. IEEE, 2017.

\bibitem{tarvainen2017mean}
Antti Tarvainen and Harri Valpola.
\newblock Mean teachers are better role models: Weight-averaged consistency targets improve semi-supervised deep learning results.
\newblock {\em Advances in neural information processing systems}, 30, 2017.

\bibitem{tasar2020standardgan}
Onur Tasar, Yuliya Tarabalka, Alain Giros, Pierre Alliez, and S{\'e}bastien Clerc.
\newblock Standardgan: Multi-source domain adaptation for semantic segmentation of very high resolution satellite images by data standardization.
\newblock In {\em Proceedings of the IEEE/CVF Conference on Computer Vision and Pattern Recognition Workshops}, pages 192--193, 2020.

\bibitem{tong2020land}
Xin-Yi Tong, Gui-Song Xia, Qikai Lu, Huanfeng Shen, Shengyang Li, Shucheng You, and Liangpei Zhang.
\newblock Land-cover classification with high-resolution remote sensing images using transferable deep models.
\newblock {\em Remote Sensing of Environment}, 237:111322, 2020.

\bibitem{tsai2018learning}
Yi-Hsuan Tsai, Wei-Chih Hung, Samuel Schulter, Kihyuk Sohn, Ming-Hsuan Yang, and Manmohan Chandraker.
\newblock Learning to adapt structured output space for semantic segmentation.
\newblock In {\em Proceedings of the IEEE conference on computer vision and pattern recognition}, pages 7472--7481, 2018.

\bibitem{tsai2019domain}
Yi-Hsuan Tsai, Kihyuk Sohn, Samuel Schulter, and Manmohan Chandraker.
\newblock Domain adaptation for structured output via discriminative patch representations.
\newblock In {\em Proceedings of the IEEE/CVF international conference on computer vision}, pages 1456--1465, 2019.

\bibitem{vu2019advent}
Tuan-Hung Vu, Himalaya Jain, Maxime Bucher, Matthieu Cord, and Patrick P{\'e}rez.
\newblock Advent: Adversarial entropy minimization for domain adaptation in semantic segmentation.
\newblock In {\em Proceedings of the IEEE/CVF conference on computer vision and pattern recognition}, pages 2517--2526, 2019.

\bibitem{Vu_2019_ICCV}
Tuan-Hung Vu, Himalaya Jain, Maxime Bucher, Matthieu Cord, and Patrick Perez.
\newblock Dada: Depth-aware domain adaptation in semantic segmentation.
\newblock In {\em Proceedings of the IEEE/CVF International Conference on Computer Vision (ICCV)}, October 2019.

\bibitem{wang2020classes}
Haoran Wang, Tong Shen, Wei Zhang, Ling-Yu Duan, and Tao Mei.
\newblock Classes matter: A fine-grained adversarial approach to cross-domain semantic segmentation.
\newblock In {\em European conference on computer vision}, pages 642--659. Springer, 2020.

\bibitem{wang2021loveda}
Junjue Wang, Zhuo Zheng, Ailong Ma, Xiaoyan Lu, and Yanfei Zhong.
\newblock Loveda: A remote sensing land-cover dataset for domain adaptive semantic segmentation.
\newblock {\em arXiv preprint arXiv:2110.08733}, 2021.

\bibitem{wang2023deep}
Xiaolei Wang, Zirong Hu, Shouhai Shi, Mei Hou, Lei Xu, and Xiang Zhang.
\newblock A deep learning method for optimizing semantic segmentation accuracy of remote sensing images based on improved unet.
\newblock {\em Scientific reports}, 13(1):7600, 2023.

\bibitem{wang2019transferable}
Ximei Wang, Ying Jin, Mingsheng Long, Jianmin Wang, and Michael~I Jordan.
\newblock Transferable normalization: Towards improving transferability of deep neural networks.
\newblock {\em Advances in neural information processing systems}, 32, 2019.

\bibitem{worldbank2024}
{World Bank}.
\newblock World development indicators 2024, 2024.

\bibitem{wrenninge2018synscapes}
Magnus Wrenninge and Jonas Unger.
\newblock Synscapes: A photorealistic synthetic dataset for street scene parsing.
\newblock {\em arXiv preprint arXiv:1810.08705}, 2018.

\bibitem{xia2023openearthmap}
Junshi Xia, Naoto Yokoya, Bruno Adriano, and Clifford Broni-Bediako.
\newblock Openearthmap: A benchmark dataset for global high-resolution land cover mapping.
\newblock In {\em Proceedings of the IEEE/CVF Winter Conference on Applications of Computer Vision}, pages 6254--6264, 2023.

\bibitem{xiao2023novel}
Ruijie Xiao, Chuan Zhong, Wankang Zeng, Ming Cheng, and Cheng Wang.
\newblock Novel convolutions for semantic segmentation of remote sensing images.
\newblock {\em IEEE Transactions on Geoscience and Remote Sensing}, 2023.

\bibitem{xie2024multimodal}
Yuxing Xie, Xiangtian Yuan, Xiao~Xiang Zhu, and Jiaojiao Tian.
\newblock Multimodal co-learning for building change detection: A domain adaptation framework using vhr images and digital surface models.
\newblock {\em IEEE Transactions on Geoscience and Remote Sensing}, 2024.

\bibitem{xiong2023benchmark}
Zhitong Xiong, Wei Huang, Jingtao Hu, and Xiao~Xiang Zhu.
\newblock The benchmark: Transferable representation learning for monocular height estimation.
\newblock {\em IEEE Transactions on Geoscience and Remote Sensing}, 2023.

\bibitem{xu2022simpl}
Yang Xu, Bohao Huang, Xiong Luo, Kyle Bradbury, and Jordan~M Malof.
\newblock Simpl: Generating synthetic overhead imagery to address custom zero-shot and few-shot detection problems.
\newblock {\em IEEE Journal of Selected Topics in Applied Earth Observations and Remote Sensing}, 15:4386--4396, 2022.

\bibitem{dfc18}
Yonghao Xu, Bo~Du, Liangpei Zhang, Daniele Cerra, Miguel Pato, Emiliano Carmona, Saurabh Prasad, Naoto Yokoya, Ronny H{\"a}nsch, and Bertrand Le~Saux.
\newblock Advanced multi-sensor optical remote sensing for urban land use and land cover classification: Outcome of the 2018 ieee grss data fusion contest.
\newblock {\em IEEE Journal of Selected Topics in Applied Earth Observations and Remote Sensing}, 12(6):1709--1724, 2019.

\bibitem{yang2021self}
Jiayu Yang, Jose~M Alvarez, and Miaomiao Liu.
\newblock Self-supervised learning of depth inference for multi-view stereo.
\newblock In {\em Proceedings of the IEEE/CVF Conference on Computer Vision and Pattern Recognition}, pages 7526--7534, 2021.

\bibitem{yang2020semantic}
Kunping Yang, Gui-Song Xia, Zicheng Liu, Bo~Du, Wen Yang, Marcello Pelillo, and Liangpei Zhang.
\newblock Semantic change detection with asymmetric siamese networks.
\newblock {\em arXiv preprint arXiv:2010.05687}, 2020.

\bibitem{yang2024depth}
Lihe Yang, Bingyi Kang, Zilong Huang, Xiaogang Xu, Jiashi Feng, and Hengshuang Zhao.
\newblock Depth anything: Unleashing the power of large-scale unlabeled data.
\newblock {\em arXiv preprint arXiv:2401.10891}, 2024.

\bibitem{yang2020fda}
Yanchao Yang and Stefano Soatto.
\newblock Fda: Fourier domain adaptation for semantic segmentation.
\newblock In {\em Proceedings of the IEEE/CVF conference on computer vision and pattern recognition}, pages 4085--4095, 2020.

\bibitem{yen20223d}
Yu-Ting Yen, Chia-Ni Lu, Wei-Chen Chiu, and Yi-Hsuan Tsai.
\newblock 3d-pl: Domain adaptive depth estimation with 3d-aware pseudo-labeling.
\newblock In {\em European Conference on Computer Vision}, pages 710--728. Springer, 2022.

\bibitem{yu2021automatic}
Dawen Yu, Shunping Ji, Jin Liu, and Shiqing Wei.
\newblock Automatic 3d building reconstruction from multi-view aerial images with deep learning.
\newblock {\em ISPRS Journal of Photogrammetry and Remote Sensing}, 171:155--170, 2021.

\bibitem{multiview3d_1}
Dawen Yu, Shunping Ji, Jin Liu, and Shiqing Wei.
\newblock Automatic 3d building reconstruction from multi-view aerial images with deep learning.
\newblock {\em ISPRS Journal of Photogrammetry and Remote Sensing}, 171:155--170, 2021.

\bibitem{yun2019cutmix}
Sangdoo Yun, Dongyoon Han, Seong~Joon Oh, Sanghyuk Chun, Junsuk Choe, and Youngjoon Yoo.
\newblock Cutmix: Regularization strategy to train strong classifiers with localizable features.
\newblock In {\em Proceedings of the IEEE/CVF international conference on computer vision}, pages 6023--6032, 2019.

\bibitem{zhang2019category}
Qiming Zhang, Jing Zhang, Wei Liu, and Dacheng Tao.
\newblock Category anchor-guided unsupervised domain adaptation for semantic segmentation.
\newblock {\em Advances in neural information processing systems}, 32, 2019.

\bibitem{zhang2003multi}
Zuxun Zhang, Jun Wu, Yong Zhang, Yongjun Zhang, and Jianqing Zhang.
\newblock Multi-view 3d city model generation with image sequences.
\newblock {\em INTERNATIONAL ARCHIVES OF PHOTOGRAMMETRY REMOTE SENSING AND SPATIAL INFORMATION SCIENCES}, 34(5/W12):351--356, 2003.

\bibitem{zhao2019geometry}
Shanshan Zhao, Huan Fu, Mingming Gong, and Dacheng Tao.
\newblock Geometry-aware symmetric domain adaptation for monocular depth estimation.
\newblock In {\em Proceedings of the IEEE/CVF Conference on Computer Vision and Pattern Recognition}, pages 9788--9798, 2019.

\bibitem{zhao2023semantic}
Wufan Zhao, Claudio Persello, and Alfred Stein.
\newblock Semantic-aware unsupervised domain adaptation for height estimation from single-view aerial images.
\newblock {\em ISPRS Journal of Photogrammetry and Remote Sensing}, 196:372--385, 2023.

\bibitem{zheng2018t2net}
Chuanxia Zheng, Tat-Jen Cham, and Jianfei Cai.
\newblock T2net: Synthetic-to-realistic translation for solving single-image depth estimation tasks.
\newblock In {\em Proceedings of the European conference on computer vision (ECCV)}, pages 767--783, 2018.

\bibitem{jonint_height2}
Zhuo Zheng, Yanfei Zhong, and Junjue Wang.
\newblock Pop-net: Encoder-dual decoder for semantic segmentation and single-view height estimation.
\newblock In {\em IGARSS 2019-2019 IEEE International Geoscience and Remote Sensing Symposium}, pages 4963--4966. IEEE, 2019.

\bibitem{zhou2023dynamic}
Zheng Zhou, Change Zheng, Xiaodong Liu, Ye~Tian, Xiaoyi Chen, Xuexue Chen, and Zixun Dong.
\newblock A dynamic effective class balanced approach for remote sensing imagery semantic segmentation of imbalanced data.
\newblock {\em Remote Sensing}, 15(7):1768, 2023.

\bibitem{zou2018unsupervised}
Yang Zou, Zhiding Yu, BVK Kumar, and Jinsong Wang.
\newblock Unsupervised domain adaptation for semantic segmentation via class-balanced self-training.
\newblock In {\em Proceedings of the European conference on computer vision (ECCV)}, pages 289--305, 2018.

\bibitem{zou2019confidence}
Yang Zou, Zhiding Yu, Xiaofeng Liu, BVK Kumar, and Jinsong Wang.
\newblock Confidence regularized self-training.
\newblock In {\em Proceedings of the IEEE/CVF international conference on computer vision}, pages 5982--5991, 2019.

\bibitem{zou2020game}
Zhengxia Zou, Tianyang Shi, Wenyuan Li, Zhou Zhang, and Zhenwei Shi.
\newblock Do game data generalize well for remote sensing image segmentation?
\newblock {\em Remote Sensing}, 12(2):275, 2020.

\end{thebibliography}

\clearpage  
{\centering
\Large \textbf{Appendix} \\  
\vspace{1em}}  
\appendix
\section{Technical Supplements}
\par In this technical supplement, we provide detailed insights and additional results to support our main paper. Section~\textcolor{blue}{\ref{sec:workflow}} outlines the generation process of the SynRS3D dataset, including the tools and plugins used. It also covers the licenses for these plugins. Section~\textcolor{blue}{\ref{sec:realdata}} discusses the data sources and licenses of the existing real-world datasets utilized in our experiments. Section~\textcolor{blue}{\ref{sec:metrics}} elaborates on the evaluation metrics for different tasks, including the proposed \(F_{1}^{HE}\) metric specifically designed for remote sensing height estimation tasks. Section~\textcolor{blue}{\ref{sec:setting}} describes the experimental setup and the selection of hyperparameters for the RS3DAda method. Section~\textcolor{blue}{\ref{sec:ablation}} presents the ablation study results and analysis for the RS3DAda method. Section~\textcolor{blue}{\ref{sec:addcombine}} provides supplementary experimental results combining SynRS3D and real data scenarios, complementing Section 5.2 of the main paper. Section~\textcolor{blue}{\ref{sec:qualitative}} showcases the qualitative visual results of RS3DAda on various tasks. Section~\textcolor{blue}{\ref{sec:bcd}} details the generation process and samples of building change detection annotations in SynRS3D, as well as the evaluation results of the source-only scenario on different real datasets. Section~\textcolor{blue}{\ref{sec:disaster}} highlights the performance of models trained on the SynRS3D dataset using RS3DAda in the critical application of disaster mapping in remote sensing.

\subsection{Detailed Generation Workflow of SynRS3D}
\label{sec:workflow}
The generation workflow of SynRS3D involves several key steps, from initializing sensor and sunlight parameters to generating the layout, geometry, and textures of the scene. This comprehensive process ensures that the generated SynRS3D mimics real-world remote sensing scenarios with high fidelity.

\noindent The main steps of the workflow are as follows:
\begin{itemize}
    \item \textbf{Initialization:} Set up the sensor and sunlight parameters using uniform and normal distributions to simulate various conditions.
    \item \textbf{Layout Generation:} Define the grid and terrain parameters to create diverse urban and natural environments.
    \item \textbf{Geometry Generation:} Specify the characteristics of roads, rivers, buildings, and vegetation, ensuring realistic representations.
    \item \textbf{Texture Generation:} Use advanced models like GPT-4~\cite{openai2023gpt4} and Stable Diffusion~\cite{rombach2022high} to generate realistic textures for different categories of land cover.
    \item \textbf{Scene Construction and Processing:} Assemble the scene with all generated components and apply textures to create visually accurate post-event and pre-event images.
    \item \textbf{Outlier Filtering:} Filter outliers based on height maps to ensure the quality and reliability of the dataset.
\end{itemize}

\noindent The detailed algorithm for this workflow is provided in Algorithm~\ref{alg:synrs3d_workflow}. The development process of SynRS3D is based on \href{https://www.blender.org/download/releases/3-4/}{Blender 3.4}, where we utilized and modified various community add-ons to facilitate the generation of SynRS3D. A comprehensive list of all the add-ons used during our development process is presented in Table~\ref{tab:addons}.

\begin{algorithm}
\caption{Generation Workflow of SynRS3D}
\label{alg:synrs3d_workflow}
\begin{adjustbox}{max width=0.95\textwidth}
\begin{minipage}{\textwidth}
\begin{algorithmic}[1]
\STATE \textbf{Initialize Parameters}
\STATE $\mathcal{S} \gets \{\text{azimuth} \sim U(a_1, a_2), \text{look\_angle} \sim \mathcal{N}(\mu_1, \sigma_1), \text{GSD} \sim \mathcal{N}(\mu_2, \sigma_2)\}$ \textcolor{blue}{\# $\mathcal{S}$: Sensor parameters}
\STATE $\mathcal{L} \gets \{\text{elevation} \sim U(e_1, e_2), \text{intensity} \sim U(i_1, i_2), \text{color} \sim [U(c_1, c_2), U(c_1, c_2), U(c_1, c_2)]\}$ \textcolor{blue}{\# $\mathcal{L}$: Sunlight parameters}

\STATE \textbf{Generate Layout}
\STATE $\mathcal{G} \gets \{\text{district\_num} \sim \text{randint}(d_1, d_2), \text{district\_size} \sim \text{randint}(s_1, s_2), \text{obj\_density} \sim U(o_1, o_2)\}$ \textcolor{blue}{\# $\mathcal{G}$: Grid parameters}
\STATE $\mathcal{T} \gets \{\text{flat\_area} \sim U(f_1, f_2), \text{mountain\_area} \sim U(m_1, m_2), \text{sea\_area} \sim U(s_1, s_2), \text{tree\_density} \sim U(t_1, t_2)\}$ \textcolor{blue}{\# $\mathcal{T}$: Terrain parameters}

\STATE \textbf{Generate Geometry}
\STATE $\mathcal{R} \gets \{\text{river\_num} \sim \text{randint}(r_1, r_2), \text{road\_num} \sim \text{randint}(r_3, r_4), \text{width} \sim U(w_1, w_2)\}$ \textcolor{blue}{\# $\mathcal{R}$: Road and River parameters}
\STATE $\mathcal{B} \gets \{\text{height} \sim U(h_1, h_2), \text{type} \in \text{select}(\text{types}), \text{roof\_angle} \sim U(ra_1, ra_2)\}$ \textcolor{blue}{\# $\mathcal{B}$: Building parameters}
\STATE $\mathcal{V} \gets \{\text{trunk} \sim \text{Sample\_Curve}(), \text{branch\_num} \sim \text{randint}(b_1, b_2), \text{leaf\_num} \sim \text{randint}(l_1, l_2)\}$ \textcolor{blue}{\# $\mathcal{V}$: Tree parameters}

\STATE \textbf{Generate Textures}
\STATE $\mathcal{C} \gets \{\text{Rangeland}, \text{Agricultural Land}, \text{Bareland}, \text{Developed Space}, \text{Road}, \text{Roof}\}$ \textcolor{blue}{\# $\mathcal{C}$: Texture categories}
\FOR{$\text{category} \in \mathcal{C}$}
    \STATE $\text{texture\_prompts} \gets \text{GPT-4}(\text{category})$
    \STATE $\text{textures}[\text{category}] \gets \text{Stable\_Diffusion}(\text{texture\_prompts})$
\ENDFOR

\STATE \textbf{Construct Scene}
\STATE $\mathcal{P}_s \gets \text{create\_scene}(\mathcal{S} \cup \mathcal{L} \cup \mathcal{G} \cup \mathcal{T} \cup \mathcal{R} \cup \mathcal{B} \cup \mathcal{V})$ \textcolor{blue}{\# $\mathcal{P}_s$: Post-event scene}
\STATE $\mathcal{Q}_s \gets \text{remove\_buildings}(\text{copy}(\mathcal{P}_s), U(rb_1, rb_2))$ \textcolor{blue}{\# $\mathcal{Q}_s$: Pre-event scene}

\STATE \textbf{Process Scene}
\STATE $\mathcal{P}_t \gets \text{apply\_textures}(\mathcal{P}_s, \text{textures})$ \textcolor{blue}{\# $\mathcal{P}_t$: Post-event scene with textures}
\STATE $\mathcal{Q}_t \gets \text{apply\_textures}(\mathcal{Q}_s, \text{textures})$ \textcolor{blue}{\# $\mathcal{Q}_t$: Pre-event scene with textures}

\STATE $\mathcal{P}_r \gets \text{render\_rgb}(\mathcal{P}_t)$ \textcolor{blue}{\# $\mathcal{P}_r$: Post-event RGB image}
\STATE $\mathcal{Q}_r \gets \text{render\_rgb}(\mathcal{Q}_t)$ \textcolor{blue}{\# $\mathcal{Q}_r$: Pre-event RGB image}

\STATE $\mathcal{P}_l \gets \text{generate\_land\_cover}(\mathcal{P}_t)$ \textcolor{blue}{\# $\mathcal{P}_l$: Post-event land cover mapping}

\STATE $\mathcal{P}_h \gets \text{generate\_height\_map}(\mathcal{P}_t)$ \textcolor{blue}{\# $\mathcal{P}_h$: Post-event height map}

\STATE $\mathcal{P}_b \gets \text{generate\_building\_mask}(\mathcal{P}_t)$ \textcolor{blue}{\# $\mathcal{P}_b$: Post-event building mask}
\STATE $\mathcal{Q}_b \gets \text{generate\_building\_mask}(\mathcal{Q}_t)$ \textcolor{blue}{\# $\mathcal{Q}_b$: Pre-event building mask}

\STATE $\mathfrak{C} \gets \text{subtract\_masks}(\mathcal{P}_b, \mathcal{Q}_b)$ \textcolor{blue}{\# $\mathfrak{C}$: Building change detection mask}

\STATE \textbf{Filter Outliers} \textcolor{blue}{\# Input: $\mathcal{P}_h$, $H_T$, $H_m$, $H_s$; Output: $\mathcal{F}_{\mathcal{P}_h}$ (Filtered height map list)}
\STATE $H_T \gets \text{threshold value}$ \textcolor{blue}{\# Set the height threshold value}
\STATE $H_m \gets \text{minimum threshold}$ \textcolor{blue}{\# Set the minimum proportion threshold}
\STATE $H_s \gets \text{steepness value}$ \textcolor{blue}{\# Set the steepness value for the sigmoid function}
\STATE $\mathcal{F}_{\mathcal{P}_h} \gets \emptyset$ \textcolor{blue}{\# Initialize the filtered height map set}
\FOR{each $n \in \mathcal{P}_h$}
    \STATE $a \gets \text{read\_image}(n)$ \textcolor{blue}{\# Read the height map as a numpy array}
    \STATE $T_p \gets \text{total\_pixels}(a)$ \textcolor{blue}{\# Calculate the total number of pixels}
    \STATE $A_t \gets \text{count\_above\_threshold}(a, H_T)$ \textcolor{blue}{\# Count the number of pixels above the threshold}
    \STATE $P_c \gets \frac{A_t}{T_p}$ \textcolor{blue}{\# Calculate the proportion of pixels above the threshold}
    \IF{$P_c \geq H_m$}
        \STATE $\mathcal{F}_{\mathcal{P}_h} \gets \mathcal{F}_{\mathcal{P}_h} \cup \{n\}$ \textcolor{blue}{\# If proportion is above minimum threshold, add to filtered list}
    \ELSE
        \STATE $Pr \gets \frac{1}{1 + e^{-H_s \cdot (P_c - H_m)}}$ \textcolor{blue}{\# Calculate the probability using a sigmoid function}
        \IF{$\text{random()} < Pr$}
            \STATE $\mathcal{F}_{\mathcal{P}_h} \gets \mathcal{F}_{\mathcal{P}_h} \cup \{n\}$ \textcolor{blue}{\# Add to filtered list based on probability}
        \ENDIF
    \ENDIF
\ENDFOR

\STATE \textbf{Output} \textcolor{blue}{\# Output: SynRS3D dataset}
\STATE $\{\mathcal{F}_{\mathcal{P}_r}, \mathcal{F}_{\mathcal{Q}_r}, \mathcal{F}_{\mathcal{P}_l}, \mathcal{F}_{\mathcal{P}_h}, \mathcal{F}_{\mathfrak{C}}\}$ \textcolor{blue}{\# $\mathcal{F}_{\mathcal{P}_r}$: Filtered post-event RGB images, $\mathcal{F}_{\mathcal{Q}_r}$: Filtered pre-event RGB images, $\mathcal{F}_{\mathcal{P}_l}$: Filtered post-event land cover mappings, $\mathcal{F}_{\mathcal{P}_h}$: Filtered post-event height maps, $\mathcal{F}_{\mathfrak{C}}$: Filtered building change detection masks}

\end{algorithmic}
\end{minipage}
\end{adjustbox}
\end{algorithm}

\begin{table}[h]
\centering
\caption{List of Blender add-ons used in the SynRS3D.}
\label{tab:addons}
\begin{adjustbox}{max width=\textwidth}
\begin{tabular}{lllll}
\toprule
\textbf{Name} & \textbf{Author} & \textbf{Version} & \textbf{License} & \textbf{URL} \\
\midrule
Realtime River Generator & specoolar & 1.1 & \href{https://support.blendermarket.com/article/49-standard-royalty-free-license}{RF} & \url{https://blendermarket.com/products/river-generator} \\
Next Street & Next Realm & 2.0 & \href{https://support.blendermarket.com/article/49-standard-royalty-free-license}{RF} & \url{https://blendermarket.com/products/next-street} \\
Objects Replacer & Georeality Design & 1.06 & \href{https://support.blendermarket.com/article/40-gnu-general-public-license}{GPL} & \url{https://blendermarket.com/products/objects-replacer/docs} \\
Albero & Greenbaburu & 0.3 & \href{https://support.blendermarket.com/article/49-standard-royalty-free-license}{RF} & \url{https://blendermarket.com/products/albero---geometry-nodes-powered-tree-generator} \\
Hira Building Generator & HiranojiStore & 0.9 & \href{https://support.blendermarket.com/article/49-standard-royalty-free-license}{RF} & \url{https://blendermarket.com/products/hira-building-generator} \\
Procedural Building Generator & Isak Waltin & 1.2.1 & \href{https://support.blendermarket.com/article/38-creative-commons-attribution-license}{CC-BY 4.0} & \url{https://blendermarket.com/products/building-gen} \\
Pro Atmo & Contrastrender & 1.0 & \href{https://support.blendermarket.com/article/40-gnu-general-public-license}{GPL} & \url{https://blendermarket.com/products/pro-atmo} \\
Modular Buildings Creator & PH Felix & 1.0 & \href{https://support.blendermarket.com/article/49-standard-royalty-free-license}{RF} & \url{https://blendermarket.com/products/modular-buildings-creator} \\
Next Trees & Next Realm & 2.0 & \href{https://support.blendermarket.com/article/49-standard-royalty-free-license}{RF} & \url{https://blendermarket.com/products/next-trees} \\
SceneCity & Arnaud & 1.9.3 & \href{https://support.blendermarket.com/article/49-standard-royalty-free-license}{RF} & \url{http://www.cgchan.com/store/scenecity} \\
Flex Road Generator & EasyNodes & 1.1.0 & \href{https://support.blendermarket.com/article/49-standard-royalty-free-license}{RF} & \url{https://www.cgtrader.com/3d-models/scripts-plugins/modelling/blender-mesh-curve-to-road} \\
Buildify & Pavel Oliva & 1.0 & \href{https://support.blendermarket.com/article/49-standard-royalty-free-license}{RF} & \url{https://paveloliva.gumroad.com/l/buildify} \\
\bottomrule
\end{tabular}
\end{adjustbox}
\end{table}

\subsection{License and Data Source of Real-World Datasets}
\label{sec:realdata}
The licenses and data sources for the real-world datasets used for evaluation and training in this work are shown in Table~\ref{table:real}. For the Potsdam, Vaihingen, GeoNRW, Nagoya, and Tokyo datasets, we used the dsm2dtm~\footnote{https://github.com/seedlit/dsm2dtm} algorithm to convert them to normalized Digital Surface Model (nDSM), since they only provide Digital Surface Model (DSM). We will release the processed real-world datasets upon acceptance, provided that the original datasets are allowed to be redistributed and are intended for non-commercial use.

\begin{table}[t]
\caption{The data source and license of real-world height estimation datasets used in this work.}
\label{table:real}
\centering
\begin{adjustbox}{width=\textwidth,center}
\begin{tabular}{c|c|c|c}
\toprule
\multicolumn{4}{c}{\textbf{Real-World Datasets}} \\
\toprule
\textbf{Types} & \textbf{Datasets} & \textbf{Data Source} & \textbf{License/Conditions of Use} \\
\midrule
\multirow{6}{*}{\begin{tabular}[c]{@{}c@{}}Target \\ Domain 1\end{tabular}} 
 & Houston \cite{dfc18} & \href{https://ieee-dataport.org/open-access/2018-ieee-grss-data-fusion-challenge-%E2%80%93-fusion-multispectral-lidar-and-hyperspectral-data}{Data Fusion Contest 2018} & \href{https://creativecommons.org/licenses/by/4.0/}{Creative Commons Attribution} \\
 & JAX \cite{dfc2019} & \href{https://ieee-dataport.org/open-access/data-fusion-contest-2019-dfc2019}{Data Fusion Contest 2019} & \href{https://creativecommons.org/licenses/by/4.0/}{Creative Commons Attribution} \\
 & OMA \cite{dfc2019} & \href{https://ieee-dataport.org/open-access/data-fusion-contest-2019-dfc2019}{Data Fusion Contest 2019} & \href{https://creativecommons.org/licenses/by/4.0/}{Creative Commons Attribution} \\
 & GeoNRW\_Urban \cite{baier2021synthesizing} & \href{https://ieee-dataport.org/open-access/geonrw}{GeoNRW} & \href{https://creativecommons.org/licenses/by/4.0/}{Creative Commons Attribution} \\
 & GeoNRW\_Rural \cite{baier2021synthesizing} & \href{https://ieee-dataport.org/open-access/geonrw}{GeoNRW} & \href{https://creativecommons.org/licenses/by/4.0/}{Creative Commons Attribution} \\
 & Potsdam \cite{rottensteiner2012isprs} & \href{https://www.isprs.org/education/benchmarks/UrbanSemLab/2d-sem-label-potsdam.aspx}{ISPRS} &  Research Purposes Only, No Redistribution \\
\midrule
\multirow{5}{*}{\begin{tabular}[c]{@{}c@{}}Target \\ Domain 2\end{tabular}} 
 & ATL \cite{ogc} & \href{https://www.drivendata.org/competitions/78/overhead-geopose-challenge/page/394/}{Overhead Geopose Challenge} & \href{https://creativecommons.org/licenses/by/4.0/}{Creative Commons Attribution} \\
 & ARG \cite{ogc} & \href{https://www.drivendata.org/competitions/78/overhead-geopose-challenge/page/394/}{Overhead Geopose Challenge} & \href{https://creativecommons.org/licenses/by/4.0/}{Creative Commons Attribution} \\
 & Nagoya \cite{aw3d2018} & NTT DATA Corporation and Inc. DigitalGlobe & End User License Agreement \\
 & Tokyo \cite{aw3d2018} & NTT DATA Corporation and Inc. DigitalGlobe & End User License Agreement \\
 & Vaihingen \cite{rottensteiner2012isprs} & \href{https://www.isprs.org/education/benchmarks/UrbanSemLab/2d-sem-label-vaihingen.aspx}{ISPRS} & Research Purposes Only, No Redistribution \\
\bottomrule
\end{tabular}
\end{adjustbox}
\end{table}

\subsection{Evaluation Metrics}
\label{sec:metrics}
We utilized several metrics to ensure a comprehensive assessment of model performance when evaluating land cover mapping and height estimation tasks. In the following parts, we provide a detailed explanation and formulation of adopted metrics.

\subsubsection{Land Cover Mapping}

\paragraph{Intersection over Union (IoU)}
Intersection over Union (IoU) is a common evaluation metric used in image segmentation tasks. It measures the overlap between the predicted segmentation and the ground truth segmentation. The IoU for a single class is defined as:

\begin{equation}
\text{IoU} = \frac{|A \cap B|}{|A \cup B|},
\end{equation}

where \(A\) is the set of predicted pixels and \(B\) is the set of ground truth pixels.

\paragraph{Mean Intersection over Union (mIoU)}
mIoU extends IoU to multiple classes by averaging the IoU values of all classes. If there are \(N\) classes, mIoU is calculated as:

\begin{equation}
\text{mIoU} = \frac{1}{N} \sum_{i=1}^{N} \text{IoU}_i,
\end{equation}

where \(\text{IoU}_i\) is the IoU for class \(i\). This metric provides a single scalar value that summarizes the segmentation performance across all classes.

\subsubsection{Height Estimation}

\paragraph{Mean Absolute Error (MAE)}
Mean Absolute Error (MAE) measures the average magnitude of the errors between the predicted heights and the true heights. Suppose the ground truth heights are \(Y\) and the predicted heights are \(\hat{Y}\),  and \(n\) is the number of samples. It is defined as:
\begin{equation}
\text{MAE} = \frac{1}{n} \sum_{i=1}^{n} |Y_i - \hat{Y}_i|.
\end{equation}

\paragraph{Root Mean Squared Error (RMSE)}
Root Mean Squared Error (RMSE) measures the square root of the average squared differences between predicted heights and actual heights. It is defined as:

\begin{equation}
\text{RMSE} = \sqrt{\frac{1}{n} \sum_{i=1}^{n} (Y_i - \hat{Y}_i)^2}.
\end{equation}

\paragraph{Accuracy Metric}
This metric, also called \(\delta\) metric from early depth estimation work \cite{eigen2014depth}, evaluates the proportion of height predictions that fall within a certain ratio of the true heights.  We use \(\delta\) to represent a maxRatio map, which is calculated as follows:

\begin{equation}
\delta = \max \left( \frac{\hat{Y}}{Y}, \frac{Y}{\hat{Y}} \right).
\label{eq:maxratio}
\end{equation}

Then, threshold values \(\eta\) are used to measure the accuracy of the height predictions, the values of \(\eta\) are usually \(1.25, 1.25^2, 1.25^3\).

\paragraph{F1 Score for Height Estimation (\(F_{1}^{HE}\))}
The \(F_{1}^{HE}\) score innovatively applies the F1 score, typically used in classification, to the regression task of height estimation. This metric emphasizes both precision and recall in estimating significant heights. The \(F_{1}^{HE}\) score balances precision and recall for height predictions above a significance threshold \(T\) (e.g., 1 meter). The maxRatio is calculated as in equation \ref{eq:maxratio}. True Positives (TP), False Positives (FP), and False Negatives (FN) are identified as follows:

\begin{align}
    \mathit{TP} &= \sum\left( (\hat{Y} > T \land Y > T) \land (\delta < \eta) \right), \\
    \mathit{FP} &= \sum\left( \hat{Y} > T \land Y \leq T \right), \\
    \mathit{FN} &= \sum\left( \hat{Y} \leq T \land Y > T \right),
\end{align}
where the values of \(\eta\) are usually \(1.25, 1.25^2, 1.25^3\). Precision, Recall, and \(F_{1}^{HE}\) are then calculated as:

\begin{align}
    \mathit{Precision} &= \frac{TP}{TP + FP}, \\
    \mathit{Recall} &= \frac{TP}{TP + FN}, \\
    \mathit{F}_{1}^{HE} &= 2 \times \frac{\mathit{Precision} \times \mathit{Recall}}{\mathit{Precision} + \mathit{Recall}}.
\end{align}

\begin{figure}[t]
\centering
\includegraphics[width=\textwidth]{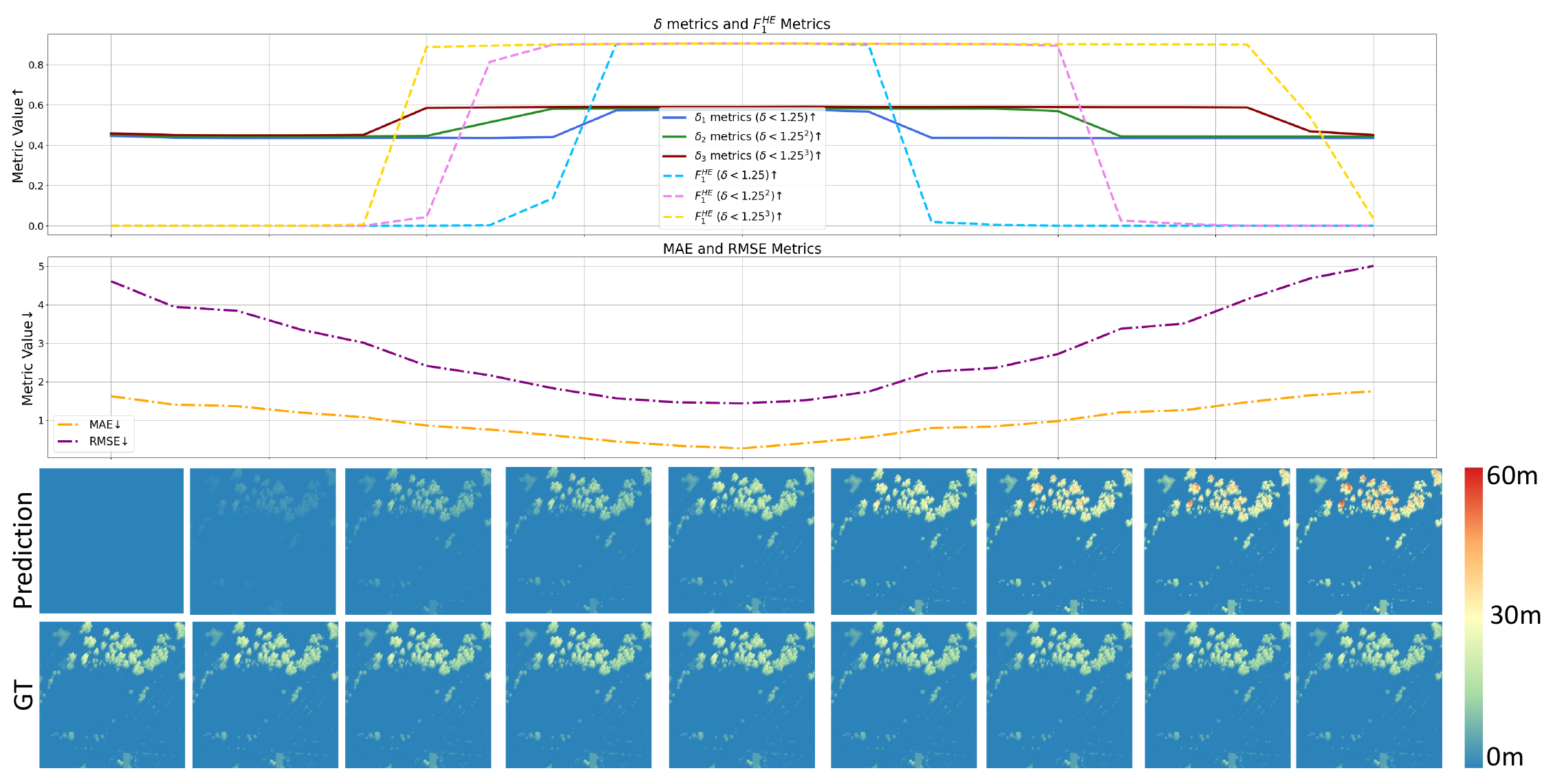}
\caption{Comparison of proposed \(F_{1}^{HE}\) metric and other metrics.}
\label{fig:delta_f1_comparison}
\end{figure}
Our motivation for proposing a new metric for height estimation arises from observing that existing metrics such as MAE, RMSE, and \(\delta\) metrics, which are derived from depth estimation tasks, do not consider the unique characteristics of height estimation in remote sensing images. Specifically, a significant portion of the remote sensing images can be occupied by ground classes, leading to an abundance of zero height values in the ground truth. This imbalance impedes the evaluation of model performance when using traditional depth estimation metrics.

As illustrated in Figure~\ref{fig:delta_f1_comparison}, when a network predicts all values as 0 meters or predicts the height of trees and buildings as twice their ground truth (30 meters to 60 meters), metrics like MAE, RMSE, and \(\delta\) still indicate highly competitive accuracy. This is not reasonable because these metrics average the correct predictions of a large number of ground pixels. However, in height estimation tasks, the accuracy of predictions for objects with height is crucial. Our proposed \(F_{1}^{HE}\) metric specifically addresses this issue by focusing on the accuracy of height predictions for objects higher than 1 meter. As shown, in both extreme cases, the F1 score is 0, reflecting the poor performance correctly. This metric better aligns with the objectives of the height estimation task. In practice, most images in height estimation datasets contain objects with heights exceeding 1 meter, so we skip the \(F_{1}^{HE}\) calculation for images that only contain ground pixels.

This comprehensive evaluation framework ensures that height estimation models are assessed on both overall error rates and the ability to accurately predict significant height values in remote sensing images.

\subsection{Experimental Setting for RS3DAda}
\label{sec:setting}
For the real-world datasets used in our experiments, we split each dataset into a 3:1 ratio for training and testing. In the RS3DAda experiments, we use random cropping of size 392 to ensure the dimensions are multiples of 14. The training batch size is set to 2, with each batch consisting of one labeled synthetic image from SynRS3D and one unlabeled image from the target domain training set. 

Additionally, in RS3DAda, the teacher model is updated using Exponential Moving Average (EMA) of the student model parameters as follows:

\begin{equation}
    \theta_t \leftarrow \alpha \theta_t + (1 - \alpha) \theta_s,
\end{equation}

where \(\theta_t\) represents the teacher model parameters, \(\theta_s\) represents the student model parameters, and \(\alpha\) is the EMA decay factor.

For detailed experimental parameters, please refer to Table~\ref{tab:expset}.

\begin{table}[t]
\centering
\caption{Settings for RS3DAda experimental hyperparameters.}
\label{tab:expset}
\begin{adjustbox}{max width=\textwidth}
\begin{threeparttable}
\begin{tabular}{lll}
\hline
\textbf{Category} & \textbf{Parameter} & \textbf{Value} \\
\hline
\multirow{2}{*}{\shortstack{Statistical Image Translation \tnote{1}}}
 & Histogram Matching (HM) & blend\_ratio\(=[0.8, 1.0]\) \\
 & Pixel Distribution Adaptation (PDA) & blend\_ratio\(=[0.8, 1.0]\), transform\_type\(=\)'standard' \\
\hline
\multirow{3}{*}{Strong Augmentation} & ClassMix & \(|\mathcal{C}|/2\) \\
 & ColorJitter \tnote{2} & $p=0.8$ \\
 & GaussianBlur \tnote{3} & $p=0.5$ \\
\hline
\multirow{2}{*}{Pseudo Label Generation} & Land Cover Confidence Threshold ($\tau$) & $0.95$ \\
 & Height Map Consistency Threshold ($\eta$) & $1.55$ \\
\hline
\multirow{9}{*}{Optimization} 
 & Optimizer & AdamW \\
 & Encoder Learning Rate (\(lr\)) & $1 \times 10^{-6}$ \\
 & Decoder Learning Rate & \(10 \times lr\)\\
 & Weight Decay & $5 \times 10^{-4}$ \\
 & Batch Size & $2$ \\
 & Iterations & $40,000$ \\
 & Warmup Steps & $1,500$ \\
 & Warmup Mode & Linear \\
 & Decay Mode & Polynomial \\
 & EMA ($\alpha$) & $0.99$ \\
\hline
\multirow{3}{*}{Loss Function} & Feature Loss Threshold ($\epsilon$) & $0.8$ \\
 & Weighting Coefficient for Target Loss ($\lambda_{target}$) & $1$ \\
 & Weighting Coefficient for Feature Loss ($\lambda_{feat}$) & $1$ \\
\hline
\end{tabular}
\begin{tablenotes}
\item[1] \url{https://albumentations.ai/docs/api_reference/augmentations/domain_adaptation/}
\item[2] \url{https://albumentations.ai/docs/api_reference/augmentations/transforms/#albumentations.augmentations.transforms.ColorJitter}
\item[3] \url{https://albumentations.ai/docs/api_reference/augmentations/blur/transforms/#albumentations.augmentations.blur.transforms.GaussianBlur}
\end{tablenotes}
\end{threeparttable}
\end{adjustbox}
\label{tab:exp_settings}
\end{table}

\subsection{Ablation Studies of RS3DAda}
\label{sec:ablation}
In this section, we mainly conduct ablation experiments on the three key modules of RS3DAda: 1) the ground mask, 2) height map consistency, and 3) feature constraints. Additionally, we performed ablation studies on different mixing strategies in the strong augmentation of the target domain and the setting of the number of categories in the land cover branch. The evaluation dataset for height estimation experiments is \textit{Target Domain 2}. For land cover mapping experiments, we employed the OEM~\cite{xia2023openearthmap} dataset for evaluation.

Table \ref{tab:ablation} presents the ablation study results for the RS3DAda method. Specifically, using DINOv2~\cite{oquab2023dinov2} and DPT~\cite{ranftl2021vision}, we find that all three modules are important for height estimation, with the ground mask and height consistency being particularly crucial. For instance, in Experiments 1 and 2, adding the ground mask reduces MAE from 6.117 to 5.652 and increases \(F_{1}^{HE}\) from 0.365 to 0.423. Adding height consistency in Experiment 3 further improves performance, reducing MAE to 5.253 and increasing \(F_{1}^{HE}\) to 0.425. The feature constraint, shown in Experiment 4, also contributes to improvements, though its impact is less significant. When all three modules are used together in Experiment 6, the best results are achieved with a MAE of 4.886, \(F_{1}^{HE}\) of 0.485, and mIoU of 48.23. For land cover mapping, height consistency is essential. Without it, the model relies on land cover confidence for height regression, which is often insufficient. This lack of confidence in the pseudo labels for the height branch hinders the improvement of the height estimation branch, subsequently affecting the land cover branch. These results indicate that both branches support each other, and inadequate learning in one branch negatively impacts the other.

Interestingly, with the weaker network combination of DeepLabv2~\cite{chen2017deeplab} and ResNet101~\cite{he2016deep} (Experiments 7-9), the feature constraint is ineffective. This is because the ImageNet-pretrained feature extractor, trained on natural images, does not generalize well to synthetic remote sensing data, unlike DINOv2's self-supervised pretraining on diverse datasets. Aligning features with the ImageNet-pretrained extractor hinders learning from synthetic data due to the significant domain gap. This demonstrates our method's effectiveness in leveraging DINOv2's features as a constraint.

\begin{table}[t!]
\centering
\caption{Ablation experiments of two types of network structures with our key modules, which were introduced in the RS3DADa section. MAE and \(F_{1}^{HE}\) serve as evaluation metrics for the height estimation tasks, and IoU is used for the land cover mapping tasks.}
\label{tab:ablation}
\begin{adjustbox}{width=0.9\textwidth,center}
\begin{tabular}{l|c|c|c|c|c|c|c}
\toprule
\# & Model & \begin{tabular}[c]{@{}c@{}}Ground\\ Mask\end{tabular} & \begin{tabular}[c]{@{}c@{}}Height\\ Consistency\end{tabular} & \begin{tabular}[c]{@{}c@{}}Feature \\ Constraint \end{tabular} & \multicolumn{2}{c|}{Height Estimation} & \begin{tabular}[c]{@{}c@{}}Land Cover Mapping\end{tabular} \\
\midrule
 & & & & MAE $\downarrow$ & \(F_{1}^{HE}\) ($\delta<1.25$) $\uparrow$ & mIoU $\uparrow$ \\
\midrule
1 & DPT+DINOv2 & $-$ & $-$ & $-$ & 6.117 & 0.365 & 42.60 \\
2 & DPT+DINOv2 & \cmark & $-$ & $-$ & 5.652 & 0.423 & 44.05 \\
3 & DPT+DINOv2 & \cmark & \cmark & $-$ & 5.253 & 0.425 & 44.75 \\
4 & DPT+DINOv2 & \cmark & $-$ & \cmark & 5.578 & 0.439 & 42.93 \\
5 & DPT+DINOv2 & $-$ & \cmark & \cmark & 5.384 & 0.461 & 46.67 \\
6 & DPT+DINOv2 & \cmark & \cmark & \cmark & \textbf{4.886} & \textbf{0.485} & \textbf{48.23} \\
\midrule
7 & DLv2+R101 & $-$ & $-$ & $-$ & 7.419 & 0.318 & 17.42 \\
8 & DLv2+R101 & \cmark & \cmark & \cmark & 6.959 & 0.316 & 18.89 \\
9 & DLv2+R101 & \cmark & \cmark & $-$ & \textbf{6.708} & \textbf{0.352} & \textbf{22.55} \\
\bottomrule
\end{tabular}
\end{adjustbox}
\end{table}

\begin{table}[ht]
\centering
\caption{Comparison of mixing strategies and number of classes.}
\label{tab:mix_class}
\begin{adjustbox}{width=0.7\textwidth,center}
\begin{tabular}{l|cc|c}
\toprule
Mix Strategy / \#Class & \multicolumn{2}{c|}{Height Estimation} & \begin{tabular}[c]{@{}c@{}}Land Cover Mapping \end{tabular} \\
\midrule
 & MAE $\downarrow$ & \(F_{1}^{HE}\) ($\delta<1.25$) $\uparrow$ & mIoU $\uparrow$ \\
\midrule
\textbf{Mix Strategy} & & & \\
CutMix~\cite{yun2019cutmix} & 4.966 & 0.475 & 47.34 \\
ClassMix~\cite{olsson2021classmix} & \textbf{4.886} & \textbf{0.485} & \textbf{48.23} \\
\midrule
\textbf{\#Classes} & & & \\
3 & 5.136 & 0.425 & - \\
8 & \textbf{4.886} & \textbf{0.485} & - \\
\bottomrule
\end{tabular}
\end{adjustbox}
\end{table}
We also explored the impact of two different mix strategies and the number of land cover classes on the RS3DAda method. As shown in Tab. \ref{tab:mix_class}, ClassMix has a slight advantage over CutMix in both tasks. Regarding the number of land cover classes, we found that using all 8 land cover classes outperforms using only 3 classes (ground, tree, building). This improvement is likely because land cover mapping, being a segmentation task, benefits from a more detailed and discrete representation of features. In contrast, height estimation, which is a regression task, relies on continuous features. By having a finer label space in the classification branch, we can better align the segmentation and regression tasks, reducing the discrepancy between them. 

\subsection{Additional Height Estimation Results in Combining SynRS3D and Real Data Scenarios}
\label{sec:addcombine}
\begin{figure}[h]
  \centering
  \includegraphics[width=\textwidth]{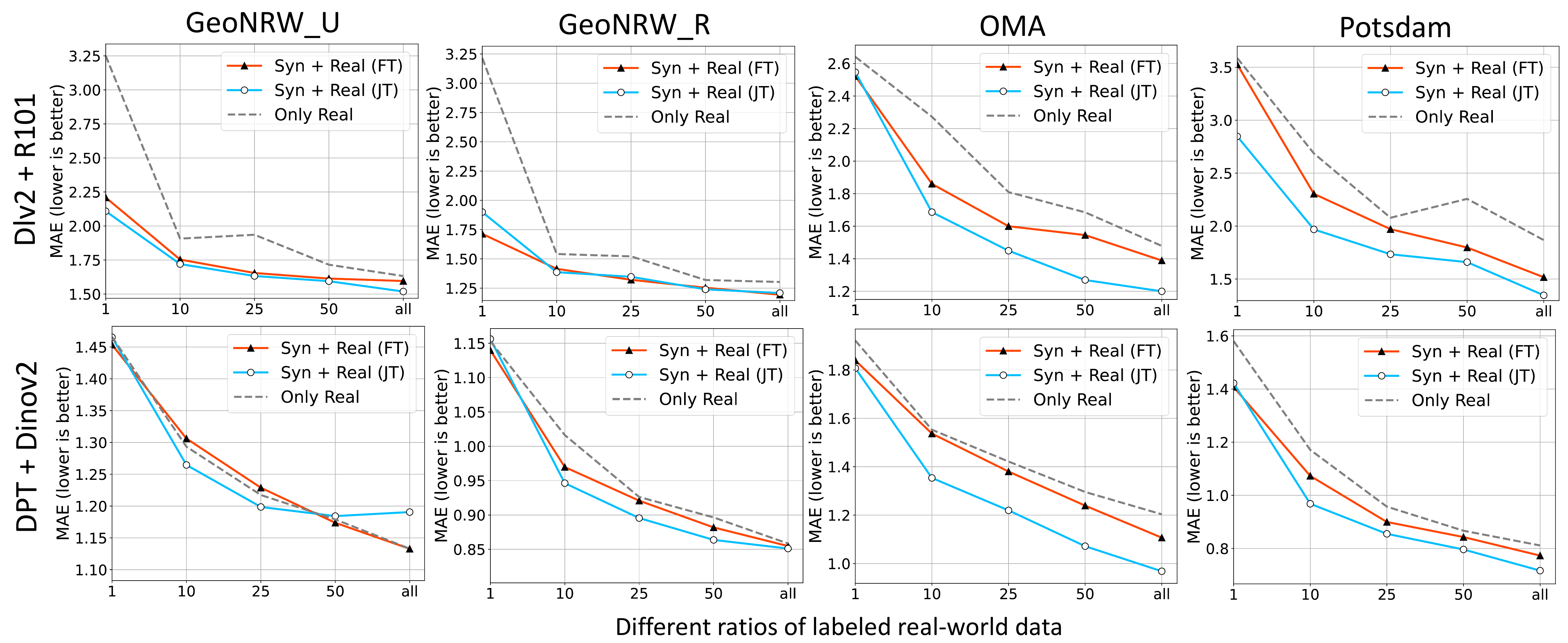}
  \caption{Additional performance evaluation on \textit{Target Domain 1} datasets of combining SynRS3D with real data on height estimation task. FT: fine-tuning on real data after pre-training on SynRS3D, JT: joint training with SynRS3D and real data.}
  \label{fig:combine1}
\end{figure}

\begin{figure}[h]
  \centering
  \includegraphics[width=\textwidth]{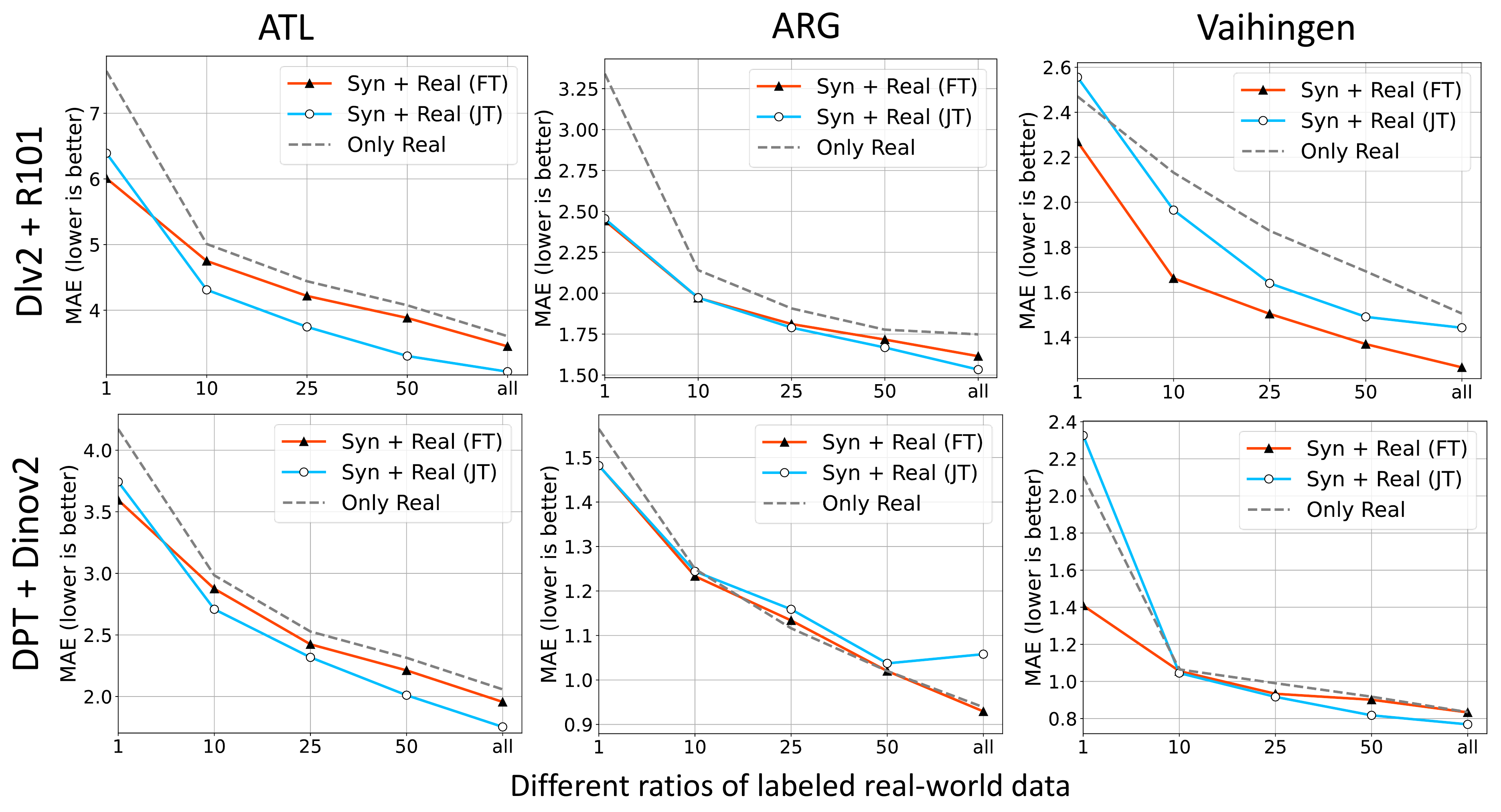}
  \caption{Additional performance evaluation on \textit{Target Domain 2} datasets of combining SynRS3D with real data on height estimation task. FT: fine-tuning on real data after pre-training on SynRS3D, JT: joint training with SynRS3D and real data.}
  \label{fig:combine2}
\end{figure}

In the Section 5.2 of the main paper, we present height estimation results for three datasets. Here, we provide the remaining results for seven additional datasets. These results further demonstrate the efficacy of combining SynRS3D with real data across different environments for fine-tuning and joint training. Figures~\ref{fig:combine1} and \ref{fig:combine2} showcase the performance across these additional datasets, following the same evaluation methodology as described in Section 5.2 of the main paper.
These extended results support the main paper's conclusions, demonstrating that both fine-tuning on real data after pre-training on SynRS3D (FT) and joint training with SynRS3D and real data (JT) significantly enhance model performance, especially when real data is limited. This underscores the importance of SynRS3D in complementing existing datasets and boosting model performance.

\subsection{Qualitative Results of RS3DAda}
\label{sec:qualitative}
\begin{figure}[htbp]
  \centering
  \includegraphics[width=\textwidth]{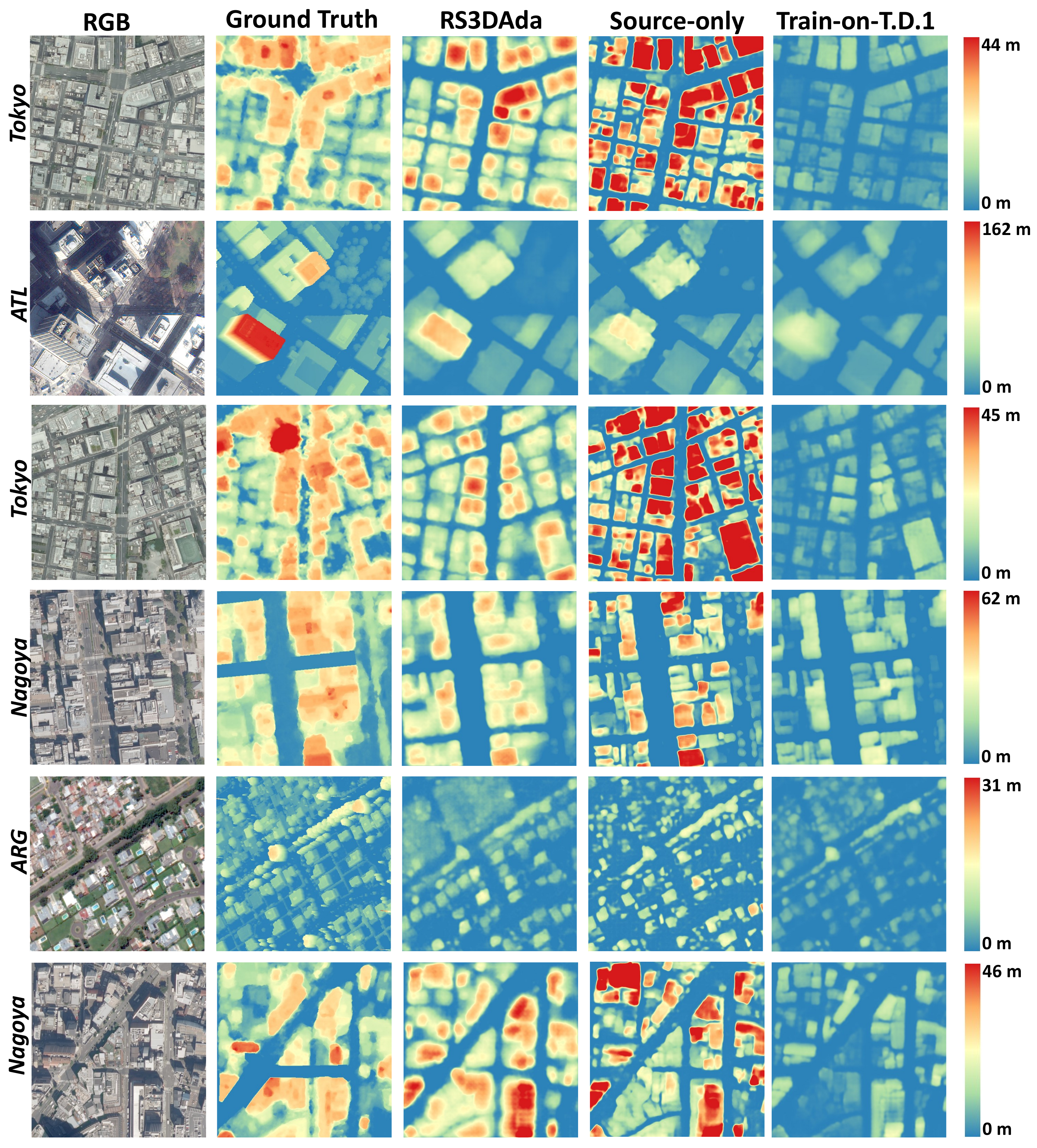}
  \caption{Qualitative results of height estimation task on \textit{Target Domain 2} using the RS3DAda model, the source-only model, and the model trained on \textit{Target Domain 1}. Satellite RGB images form Tokyo and Nagoya: © 2018 NTT DATA Corporation and Inc. DigitalGlobe.}
  \label{fig:qualitive_he}
\end{figure}

\begin{figure}[htbp]
  \centering
  \includegraphics[width=\textwidth]{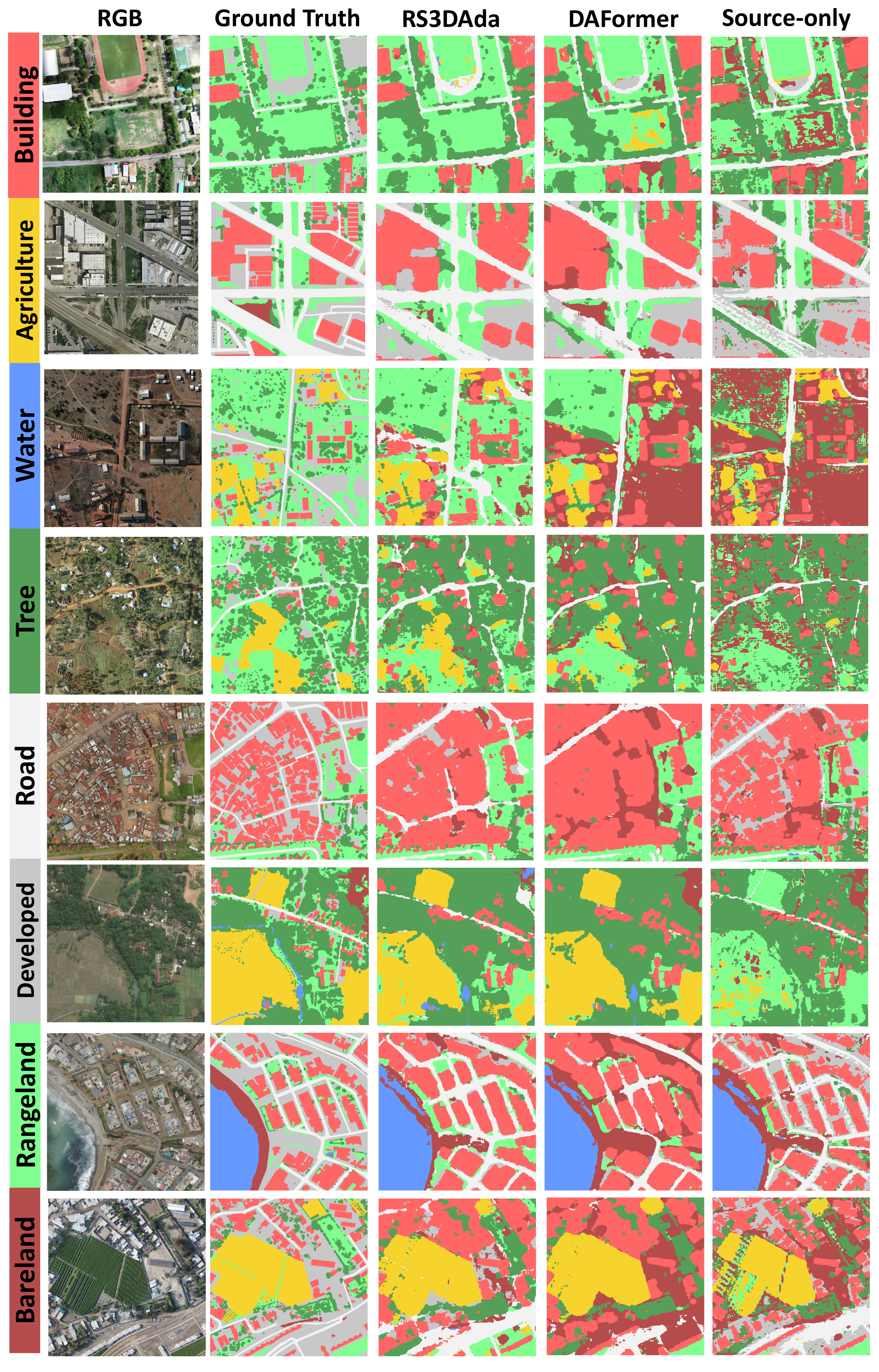}
  \caption{Qualitative results of land cover mapping task on OEM dataset using the RS3DAda model, the source-only model, and DAFormer.}
  \label{fig:qualitive_ss}
\end{figure}

\begin{figure}[htbp]
  \centering
  \includegraphics[width=0.95\textwidth]{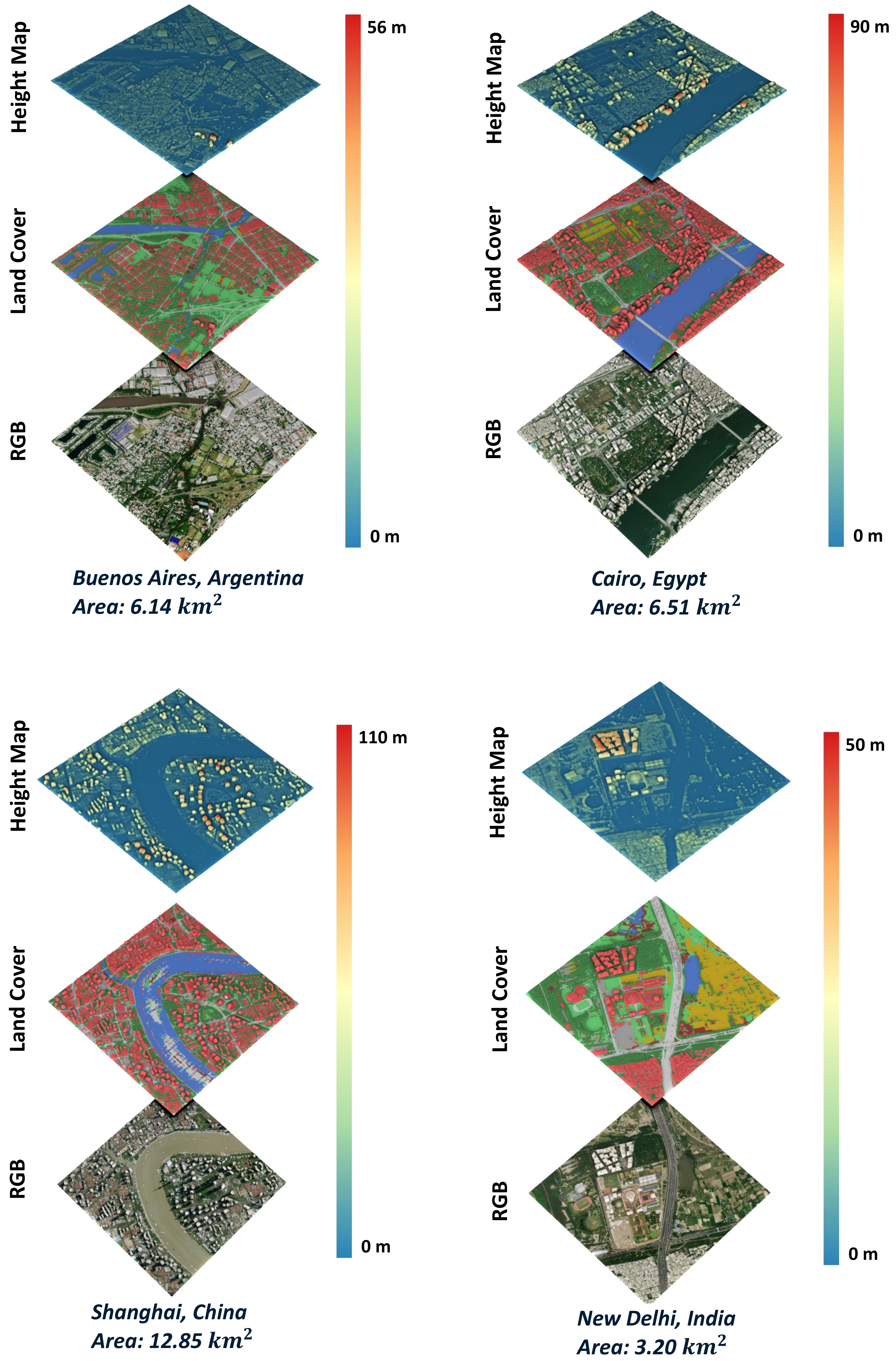}
  \caption{3D visualization outcomes from real-world monocular RS images, which uses the model trained on SynRS3D dataset with proposed RS3DAda method. RGB satellite images of Buenos Aires and New Delhi: © HERE WeGo Satellite. RGB satellite images of Cairo and Shanghai: © Being Satellite.}
  \label{fig:add_recon}
\end{figure}

Figure~\ref{fig:qualitive_he} shows the qualitative results for the height estimation task. We can observe that the height predictions from the RS3DAda model are closer to the ground truth and have more complete edges. In contrast, the source-only model tends to overestimate height values and produces more incomplete edges. Although the model trained on \textit{Target Domain 1} uses real data, it struggles to generalize to \textit{Target Domain 2} due to its training data being limited to commonly available public datasets from European and American regions, which are unbalanced. As shown, its predicted heights are often underestimated.
Figure~\ref{fig:qualitive_ss} presents the qualitative results for the land cover mapping task. The RS3DAda model demonstrates exceptional performance in categories such as agricultural land, rangeland, and bare land, which aligns with our quantitative experimental results. However, it has some limitations in categories like roads and developed space, indicating that there is still significant room for improvement in domain adaptation research for the SynRS3D dataset in the area of land cover mapping. This marks the first time in the field of remote sensing that synthetic data alone can achieve a high level of visual interpretation consistency with the ground truth. We hope that the RS3DAda method and the SynRS3D dataset can serve as benchmarks to further advance research in this direction.
Figure~\ref{fig:add_recon} shows additional 3D reconstruction results in developing countries. These results are derived from using models trained on SynRS3D with RS3DAda to infer monocular satellite image tiles from Bing Satellite\footnote{\url{https://www.bing.com/maps}} and HereWeGo Satellite\footnote{\url{https://wego.here.com}}. These 3D reconstruction areas cover between 3.2 square kilometers and 12.85 square kilometers, with a ground sample distance (GSD) of 0.35 meters.
\subsection{Building Change Detection}
\label{sec:bcd}
\begin{figure}[htbp]
\centering
\includegraphics[width=\textwidth]{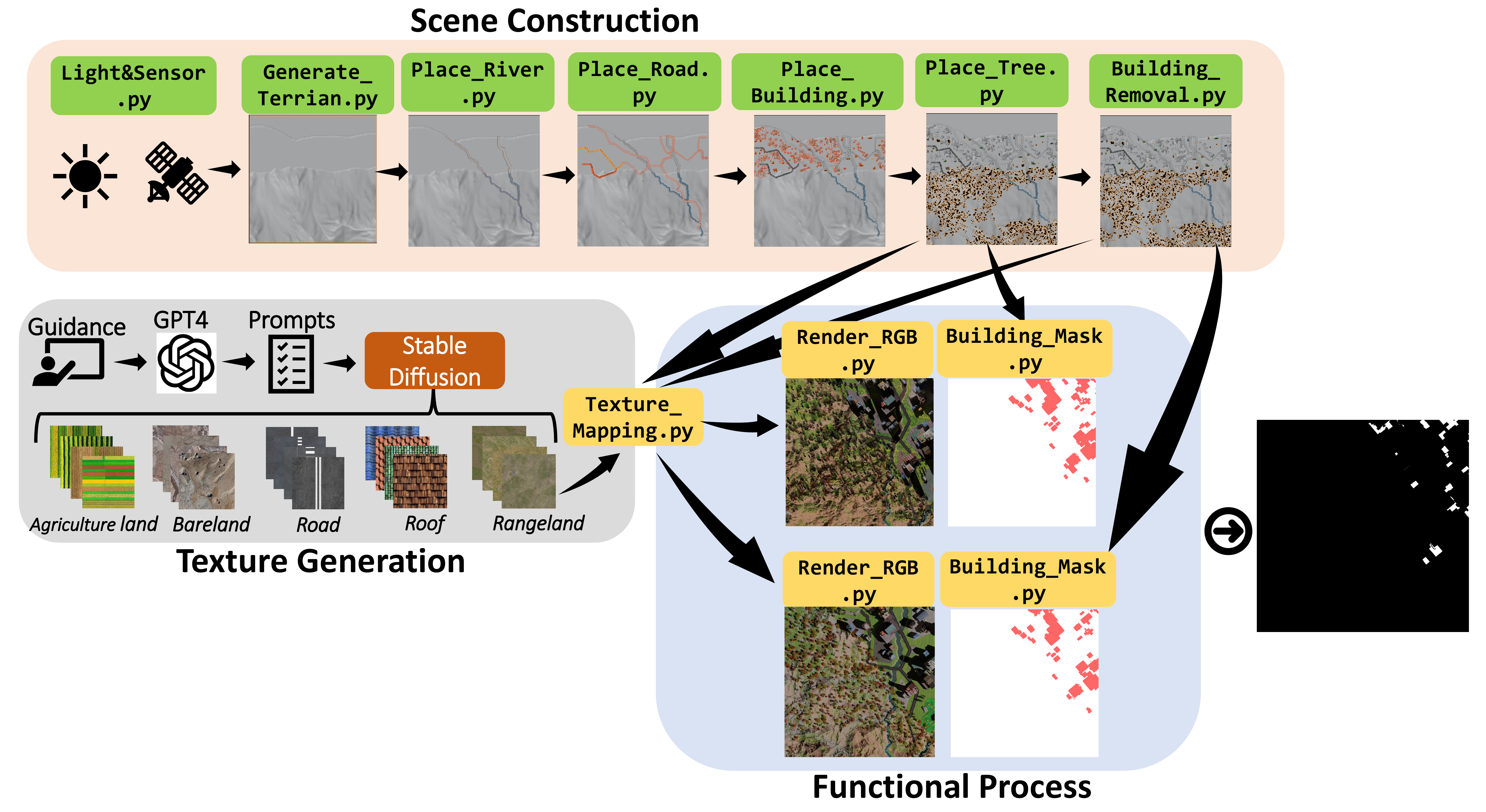}
\caption{Generation workflow of building change detection mask.}
\label{fig:cdworkflow}
\end{figure}
\begin{figure}[t]
\centering
\includegraphics[width=\textwidth]{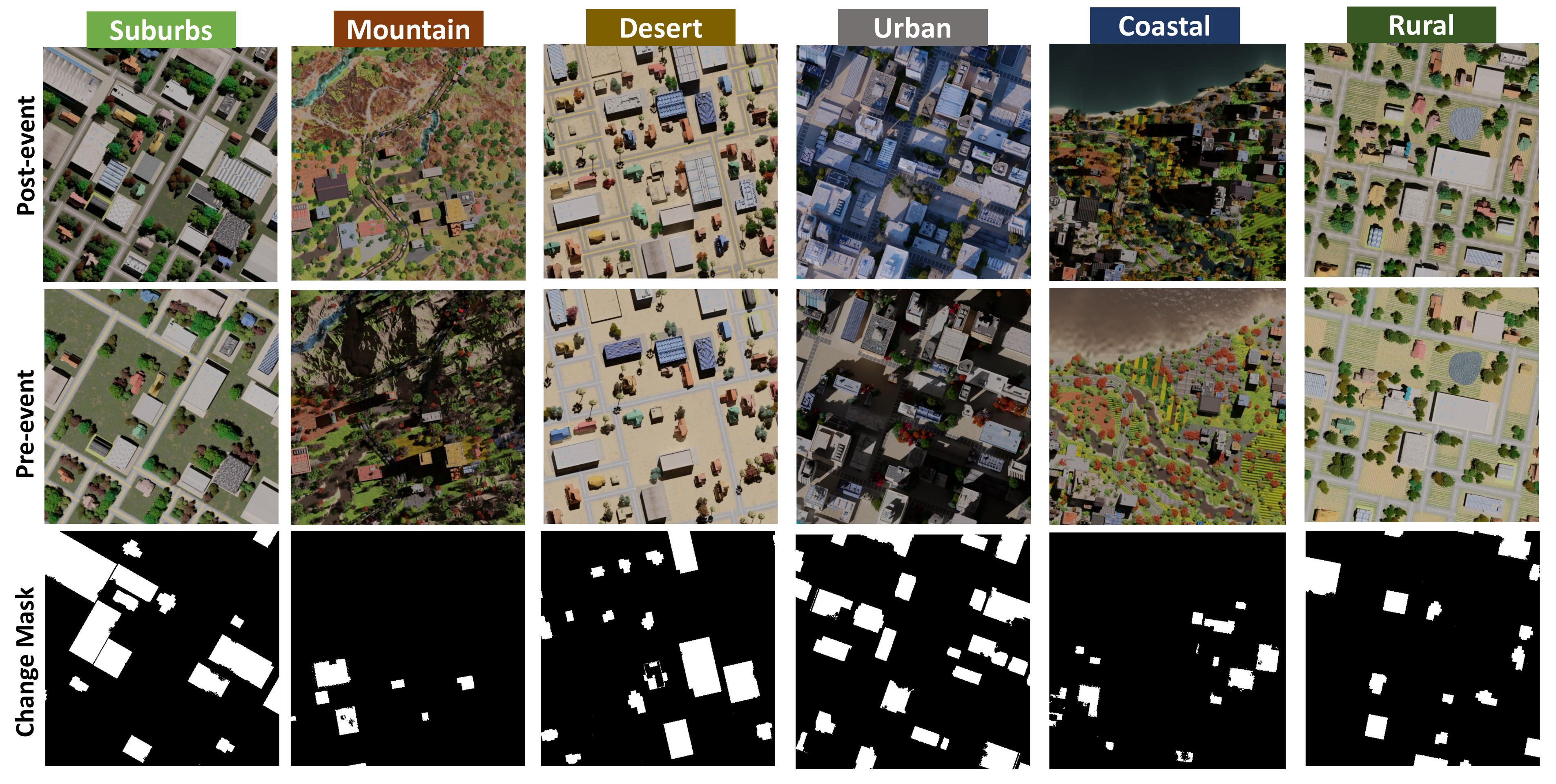}
\caption{Examples of building change detection task in SynRS3D.}
\label{fig:cdexamples}
\end{figure}

SynRS3D provides 8-class land cover mapping annotations, accurate height maps, and binary masks specifically designed for RS building change detection tasks. The image and mask generation process is illustrated in Figure~\ref{fig:cdworkflow}. For the synthesized scenes, an additional step is included: a certain proportion of buildings are randomly removed, and all geometries in the scene are retextured. Subsequently, post-event and pre-event RGB images, along with building masks, are rendered. By subtracting the two masks, the final building change mask is obtained. Figure~\ref{fig:cdexamples} exhibits examples of change detection in six styles within SynRS3D.

To validate the effectiveness of SynRS3D in change detection tasks, we conducted experiments in a source-only scenario, where models were trained only on synthetic data and tested directly on real-world datasets. We compared our results with the models trained on two other advanced synthetic datasets, SMARS~\cite{reyes20232d} and SyntheWorld~\cite{song2024syntheworld}, that include labels for the RS building change detection task. For real-world datasets, we used commonly utilized datasets in change detection tasks: WHU-CD~\cite{ji2018fully}, LEVIR-CD+~\cite{chen2020spatial}, and SECOND~\cite{yang2020semantic}. 

The WHU-CD dataset, a subset of the WHU Building dataset, focuses on building change detection with aerial images from Christchurch, New Zealand, captured in April 2012 and 2016 at 0.3 meters/pixel resolution. Covering 20.5 km\textsuperscript{2}, the dataset documents significant urban development, with buildings increasing from 12,796 to 16,077 over four years. LEVIR-CD+ is an advanced building CD dataset comprising 985 pairs of high-resolution (0.5 meters/pixel) images, documenting changes over 5 to 14 years and featuring various building types. It includes 31,333 instances of building changes, making it a valuable benchmark for CD methodologies. The SECOND dataset consists of 4,662 pairs of 512$\times$512 aerial images (0.5-3 meters/pixel) annotated for land cover change detection in cities like Hangzhou, Chengdu, and Shanghai, but in our experiments, we only use its building change mask. These datasets were split into training and testing sets in a 3:1 ratio, with a training size of 256$\times$256 pixels.

We employed four change detection frameworks for evaluating SMAR, SyntheWorld, and SynRS3D, including the CNN-based DTCDSCN~\cite{liu2020building}, the transformer-based ChangeFormer~\cite{bandara2022transformer}, and the current state-of-the-art Mamba-based method, ChangeMamba~\cite{chen2024changemamba}. Notably, due to our empirical findings of the strong potential of DINOV2~\cite{oquab2023dinov2} pre-trained networks on synthetic data in both land cover mapping and change detection tasks, we implemented a framework combining the DINOV2 encoder with the ChangeMamba decoder for change detection on synthetic data, which we named DINOMamba. For synthetic datasets, we use a batch size of 2, and for real data, we use a batch size of 16. The optimizer used is AdamW, with a learning rate of 1e-5 for DinoMamba, while all other methods use a learning rate of 1e-4. All models are trained for 40,000 iterations on a single Tesla A100. The evaluation metrics used are IoU and F1.

\begin{table}[ht]
\caption{Peformance evaluation of building change detection task on WHU-CD~\cite{ji2018fully} dataset.}
\label{tab:whu}
\centering
\begin{adjustbox}{width=0.8\textwidth,center}
\begin{tabular}{l|cc|cc|cc|cc}
\toprule
\multirow{2}{*}{Train on} & \multicolumn{2}{c|}{DTCDSCN~\cite{liu2020building}} & \multicolumn{2}{c|}{ChangeFormer~\cite{bandara2022transformer}} & \multicolumn{2}{c|}{ChangeMamba~\cite{chen2024changemamba}} & \multicolumn{2}{c}{DinoMamba} \\
\cmidrule(lr){2-9}
 & IoU & F1 & IoU & F1 & IoU & F1 & IoU & F1 \\
\midrule
SMARS~\cite{reyes20232d} & 26.84 & 42.55 & 18.67 & 31.88 & 42.50 & 59.63 & 48.11 & 64.87 \\
SyntheWorld~\cite{song2024syntheworld} & 30.17 & 46.53 & \textbf{41.73} & \textbf{58.87} & 47.26 & 64.10 & 54.20 & 70.14 \\
\rowcolor{blue!20}SynRS3D &  \textbf{33.09}& \textbf{49.84} & 35.00 & 51.94 & \textbf{52.94} & \textbf{69.08} & \textbf{61.60} & \textbf{76.00}  \\
\midrule
\rowcolor{gray!20}\textcolor{gray!80}{Real} & \textcolor{gray!80}{58.31} & \textcolor{gray!80}{73.67} & \textcolor{gray!80}{79.98} & \textcolor{gray!80}{88.88} & \textcolor{gray!80}{88.44} & \textcolor{gray!80}{93.87} &  \textcolor{gray!80}{87.57} & \textcolor{gray!80}{93.38} \\
\bottomrule
\end{tabular}
\end{adjustbox}
\end{table}

\begin{table}[ht]
\caption{Peformance evaluation of building change detection task on LEVIR-CD+~\cite{chen2020spatial} dataset.}
\label{tab:levir}
\centering
\begin{adjustbox}{width=0.8\textwidth,center}
\begin{tabular}{l|cc|cc|cc|cc}
\toprule
\multirow{2}{*}{Train on} & \multicolumn{2}{c|}{DTCDSCN~\cite{liu2020building}} & \multicolumn{2}{c|}{ChangeFormer~\cite{bandara2022transformer}} & \multicolumn{2}{c|}{ChangeMamba~\cite{chen2024changemamba}} & \multicolumn{2}{c}{DinoMamba} \\
\cmidrule(lr){2-9}
 & IoU & F1 & IoU & F1 & IoU & F1 & IoU & F1 \\
\midrule
SMARS~\cite{reyes20232d} & 11.70 & 21.53 & 15.67 & 27.58 & 27.50 & 42.50 & 30.85 & 47.31 \\
SyntheWorld~\cite{song2024syntheworld} & 21.16 & 35.28 & 23.31 & 38.12 & 28.28 & 44.30 & 48.78 & 65.46 \\
\rowcolor{blue!20}SynRS3D &  \textbf{25.82}& \textbf{41.30} & \textbf{23.33} & \textbf{38.14} & \textbf{30.39} & \textbf{46.78} & \textbf{49.63} & \textbf{66.23}  \\
\rowcolor{gray!20}\textcolor{gray!80}{Real} & \textcolor{gray!80}{63.44} & \textcolor{gray!80}{77.63} & \textcolor{gray!80}{67.48} & \textcolor{gray!80}{80.58} & \textcolor{gray!80}{77.39} & \textcolor{gray!80}{87.25} &  \textcolor{gray!80}{74.12} & \textcolor{gray!80}{85.14} \\
\bottomrule
\end{tabular}
\end{adjustbox}
\end{table}

\begin{table}[h!]
\caption{Peformance evaluation of building change detection task on the SECOND~\cite{yang2020semantic} dataset.}
\label{tab:second}
\centering
\begin{adjustbox}{width=0.8\textwidth,center}
\begin{tabular}{l|cc|cc|cc|cc}
\toprule
\multirow{2}{*}{Train on} & \multicolumn{2}{c|}{DTCDSCN~\cite{liu2020building}} & \multicolumn{2}{c|}{ChangeFormer~\cite{bandara2022transformer}} & \multicolumn{2}{c|}{ChangeMamba~\cite{chen2024changemamba}} & \multicolumn{2}{c}{DinoMamba} \\
\cmidrule(lr){2-9}
 & IoU & F1 & IoU & F1 & IoU & F1 & IoU & F1 \\
\midrule
SMARS~\cite{reyes20232d} & 17.26 & 29.88 & 23.30 & 38.09 & 29.85 & 46.15 & 35.20 & 51.07 \\
SyntheWorld~\cite{song2024syntheworld} & 21.00 & 35.07 & 26.44 & 42.06 & 27.23 & 43.02 & 37.61 & 54.71 \\
\rowcolor{blue!20}SynRS3D &  \textbf{33.52} & \textbf{50.32} & \textbf{31.36} & \textbf{47.90} & \textbf{38.88} & \textbf{56.02} & \textbf{39.18} & \textbf{56.33}  \\
\rowcolor{gray!20}\textcolor{gray!80}{Real} & \textcolor{gray!80}{58.78} & \textcolor{gray!80}{74.04} & \textcolor{gray!80}{60.08} & \textcolor{gray!80}{75.06} & \textcolor{gray!80}{67.61} & \textcolor{gray!80}{80.68} &  \textcolor{gray!80}{67.65} & \textcolor{gray!80}{80.71} \\
\bottomrule
\end{tabular}
\end{adjustbox}
\end{table}
Tables~\ref{tab:whu},~\ref{tab:levir},~\ref{tab:second} present our experimental results, showing that the combination of SynRS3D and DINOMamba achieved F1 scores of 76.00, 66.23, and 56.33 on WHU, LEVIR-CD+, and SECOND respectively. Although there is still a gap compared to the Oracle model trained on real-world data, our dataset significantly boosts models' performances compared with the other two synthetic datasets. We have established a benchmark based on SynRS3D and advanced change detection networks, hoping to further promote the development of RS change detection using synthetic data.

\subsection{Disaster Mapping Study Cases}
\label{sec:disaster}
The models trained on the SynRS3D dataset using the RS3DAda method can be utilized for various remote sensing downstream applications. We explored their potential in disaster mapping applications.

In February 2023, a devastating earthquake struck southeastern Turkey, primarily affecting the Kahramanmaraş region. This earthquake, with a magnitude of 7.8, caused widespread destruction, resulting in over 45,000 deaths, thousands of injuries, and massive displacement of residents. The economic losses were estimated to be in the billions of dollars. Rescue operations were carried out by both national and international teams, working tirelessly to save lives and provide aid to the affected population. Similarly, in August 2023, Hawaii experienced severe wildfires, particularly affecting the island of Maui. These wildfires, exacerbated by dry conditions and strong winds, led to extensive destruction of homes, infrastructure, and natural landscapes. The fires caused significant economic losses, displacing many residents and leading to casualties. The coordinated efforts of local authorities and fire departments, along with support from federal agencies, were crucial in controlling the fires and assisting those affected.
\begin{figure}[h!]
  \centering
  \includegraphics[width=\textwidth]{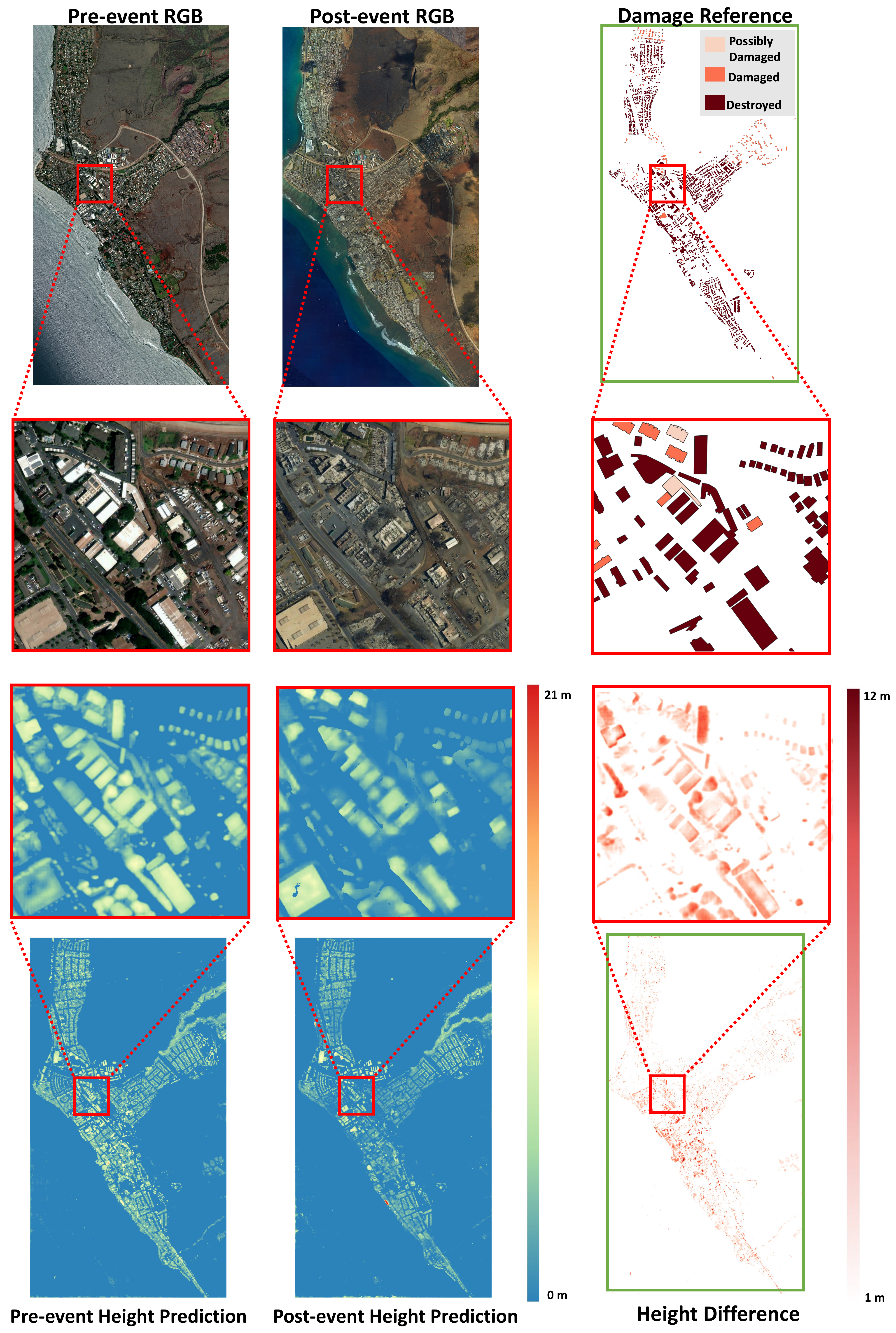}
  \caption{Study case of 2023 Hawaii-Maui wildfire. RGB satellite images of pre-event: © Being Satellite. RGB satellite images of post-event: © Google Satellite.}
  \label{fig:hawaii}
\end{figure}
\begin{figure}[h!]
  \centering
  \includegraphics[width=\textwidth]{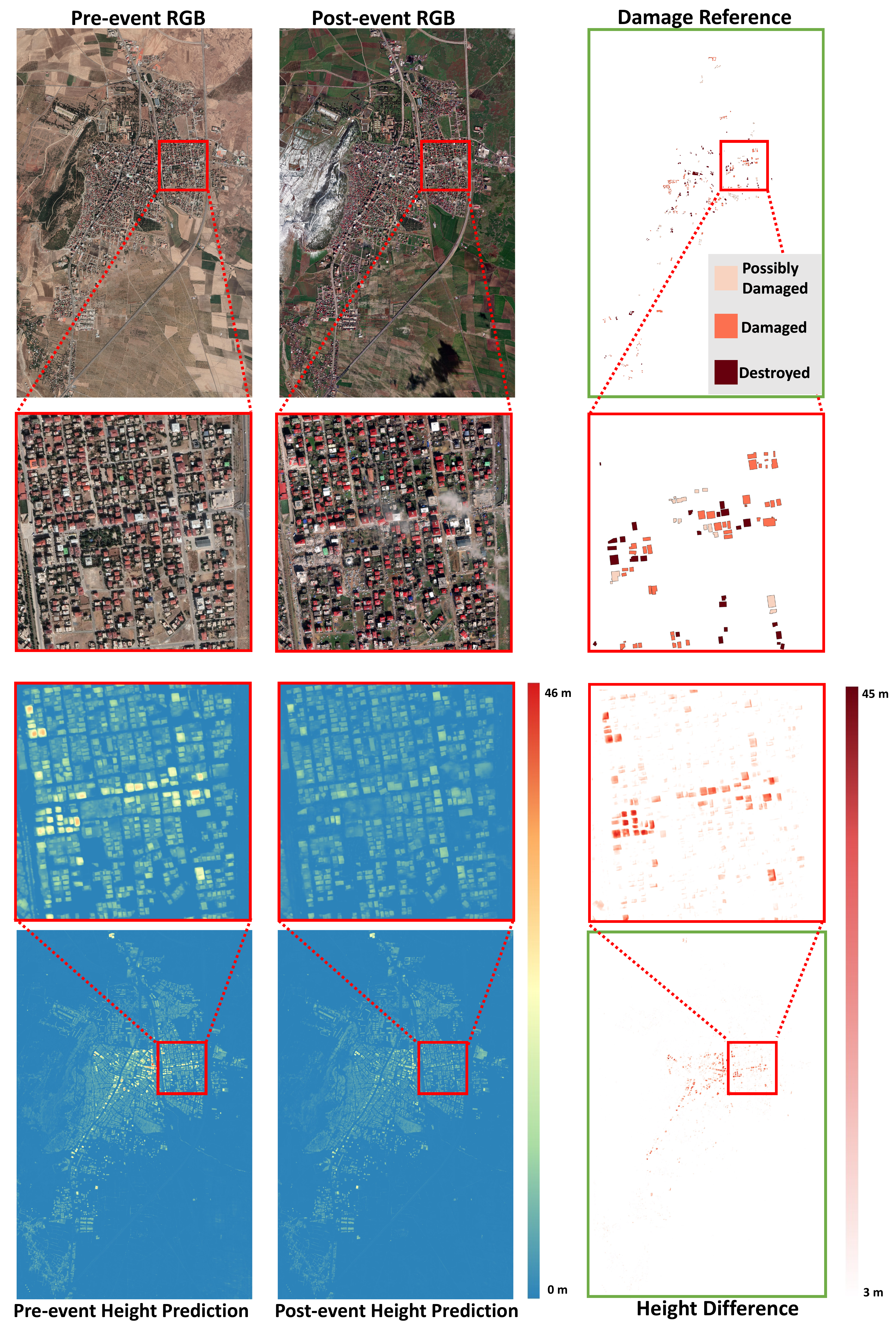}
  \caption{Study case of 2023 Turkey–Syria earthquakes. RGB satellite images: © 2023 CNES/Airbus, Maxar Technologies.}
  \label{fig:turkey}
\end{figure}
To assess the impact of these disasters, we used the height estimation branch of RS3DAda to infer pre- and post-event remote sensing images. By simply subtracting the predicted height maps of the post-event from the pre-event, we obtained a Height Difference map. This map was filtered using a threshold: 3 meters for the earthquake example (indicating that buildings severely damaged in the earthquake would collapse, resulting in a significant height reduction) and 1 meter for the wildfire example (assuming that changes exceeding 1 meter indicate damage in the fire). Figure~\ref{fig:turkey} presents the study case for the Turkey earthquake, and Figure~\ref{fig:hawaii} shows the study case for the Hawaii wildfires.

This simple method allowed us to roughly delineate the affected areas and assess the damage severity based on height differences. Although not entirely precise, this approach represents a significant success in applying models trained solely on synthetic data to real-world scenarios. We believe in the potential of RS3DAda and SynRS3D in this research domain and look forward to more applications and studies in the future.

\end{document}